\documentclass[10pt,twocolumn,letterpaper]{article}

\usepackage{wacv}

\makeatletter
\@namedef{ver@everyshi.sty}{}
\makeatother
\usepackage{tikz}

\usepackage{tcolorbox}

\usepackage{times}
\usepackage{epsfig}
\usepackage{graphicx}
\usepackage{amsmath}
\usepackage{amssymb}
\usepackage{url}            
\usepackage{booktabs}       
\usepackage{amsfonts}       
\usepackage{nicefrac}       
\usepackage{microtype}      

\usepackage[export]{adjustbox}
\usepackage{subfiles}
\usepackage{graphicx}
\usepackage{epsfig}
\usepackage{subfig}
\usepackage{amsmath}
\usepackage{amssymb} 
\usepackage{booktabs}
\usepackage[figuresright]{rotating}
\usepackage{enumitem}
\usepackage{multirow}
\usepackage{subfig}
\usepackage{hhline}

\usepackage{adjustbox}
\usepackage{xspace}
\usepackage{amssymb}
\usepackage{algorithm}
\usepackage[noend]{algpseudocode}
\usepackage{xcolor}
\usepackage{xspace}
\usepackage{cite}
\usepackage{amsthm}
\usepackage{float}

\usepackage{xpatch}

\usepackage{enumitem}
\usepackage[accsupp]{axessibility}

\definecolor{beaublue}{rgb}{0.84, 0.9, 0.95}
\definecolor{blackish}{rgb}{0.2, 0.2, 0.2}

\definecolor{beaublue2}{rgb}{0.84, 0.9, 0.95}
\definecolor{blackish2}{rgb}{0.2, 0.2, 0.2}

\makeatletter
\newcommand\fs@nobottomruled{\def\@fs@cfont{\bfseries}\let\@fs@capt\floatc@ruled
  \def\@fs@pre{}
  \def\@fs@post{}
  \def\@fs@mid{\kern2pt\hrule\kern2pt}%
  \let\@fs@iftopcapt\iftrue}
\makeatother

\makeatletter
\DeclareRobustCommand\bmvaOneDot{\futurelet\@let@token\bmv@onedotaux}
\def\bmv@onedotaux{\ifx\@let@token.\else.\null\fi\xspace}
\makeatother
\def\eg{\emph{e.g}\bmvaOneDot}

\def\etal{\emph{et al}\bmvaOneDot}
\def\etc{\emph{etc}\bmvaOneDot}
\def\ie{\emph{ie}\bmvaOneDot}

\DeclareMathOperator*{\argmax}{arg\,max}
\DeclareMathOperator*{\argmin}{arg\,min}

\def\Vec#1{{\boldsymbol{#1}}}

\usepackage[pagebackref=true,breaklinks=true,colorlinks,bookmarks=false]{hyperref}

\def\RED#1{{}{\color{red}{#1}}}
\def\BLUE#1{{}{\color{blue}{#1}}}
\wacvfinalcopy

\def\wacvPaperID{****} 

\ifwacvfinal
\fi

\ifwacvfinal
\usepackage[breaklinks=true,bookmarks=false]{hyperref}
\else
\usepackage[pagebackref=true,breaklinks=true,colorlinks,bookmarks=false]{hyperref}
\fi

\ifwacvfinal
\fi

\begin{document}

\title{
{Towards a Robust Differentiable Architecture Search under Label Noise}}

\author{%
\vspace{0.3cm}
  Christian Simon$^{\dagger, \S}$, \quad Piotr Koniusz$^{\S,\dagger}$, \quad Lars Petersson$^{\S,\dagger}$, \quad Yan Han$^{\dagger,\S}$, \quad Mehrtash Harandi$^{\clubsuit, \S}$\\\vspace{0.3cm}
  $^{\dagger}$The Australian National University \quad $^{\clubsuit}$Monash University \quad
   $^\S$Data61-CSIRO\\
  firstname.lastname\texttt{@\{anu.edu.au,monash.edu,data61.csiro.au\}} \\
}

\maketitle

\begin{abstract}
 
Neural Architecture Search (NAS) is the game changer in designing robust neural architectures.
Architectures designed by NAS outperform or compete with the best manual network designs in terms of accuracy, size, memory footprint and FLOPs. That said, previous studies focus on developing NAS algorithms for clean high quality data, a restrictive and somewhat unrealistic assumption. In this paper, focusing on the differentiable NAS algorithms, we show that vanilla NAS algorithms suffer from a performance loss if class labels are noisy. To combat this issue, we make use of the principle of  information bottleneck as a regularizer. This leads us to develop a noise injecting operation that is included during the learning process, preventing the network from learning from noisy samples. Our empirical evaluations show that the noise injecting operation does not degrade the performance of the NAS algorithm if the data is indeed clean. In contrast, if the data is noisy, the architecture learned by our algorithm comfortably outperforms algorithms specifically equipped with sophisticated mechanisms to learn in the presence of label noise. In contrast to many algorithms designed to work in the presence of noisy labels, prior knowledge about the properties of the noise and its characteristics are not required for our algorithm.

\end{abstract}


\vspace{-0.5cm}
\section{Introduction}
\label{sec:introduction}

To avoid exhausting engineering, Neural Architecture Search (NAS) has emerged as a leading mechanism for automatic design and wiring of neural networks.  
NAS has been successfully moving forward with diverse approaches to achieve a robust automatic architecture search \eg, evolution-based NAS~\cite{elsken2018efficient,esteban2017EvolLarge,stanley2019neuroevol,real2019regularized},  optimization-based NAS~\cite{Liu2018Darts, Dong2019GoruGpuHrs,luo2018neuralarchopt,saxena2016convolutional,veniat2018learning}, and Reinforcement Learning (RL) based NAS~\cite{Zoph2017NASRL,zoph2018learningscalable,pham2018ENAS}. Specifically, a gradient-based method with a continuous architecture space called Differentiable ARchiTecture Search (DARTS) has attracted significant attention in NAS because of a reduced cost and complexity of searching for high performance architectures. In this paper, we go beyond merely learning with NAS under regular assumptions of clean labels. 

Supervised learning with neural networks often leads to a performance degradation due to overfitting, especially in the presence of label noise which often emerges due to data corruption and/or human annotation errors especially prominent in large scale datasets. As a result, neural networks fail to generalize well to previously unseen data and achieve suboptimal classification results. Given the importance of such problems, existing state-of-the-art methods are specifically designed to deal with the data noise by correcting labels~\cite{vahdat2017toward}, employing dedicated loss functions~\cite{arazo2019unsupervised,Wang2019SymLoss,natarajan2013learning,zhang2018generalized}, reweighting samples~\cite{ren2018learning,jiang2018mentornet}, selecting samples~\cite{Han2018Co}, and modeling a transition matrix~\cite{xia2019anchor,patrini2017making,tanno2019learning}. However, designing robust neural networks that mitigate overfitting due to label noise is still unexplored. To this end, we propose a structural approach, that is a method that requires no explicit changes to neither the loss functions nor the input samples nor the final outputs (predictions) but the neural network structure. Moreover, our approach does not require specific assumptions on the amount or the type of label noise.

\begin{figure*}[t]
\vspace{-0.6cm}
    \centering
 \subfloat[]{
        \includegraphics[width=0.242\textwidth]{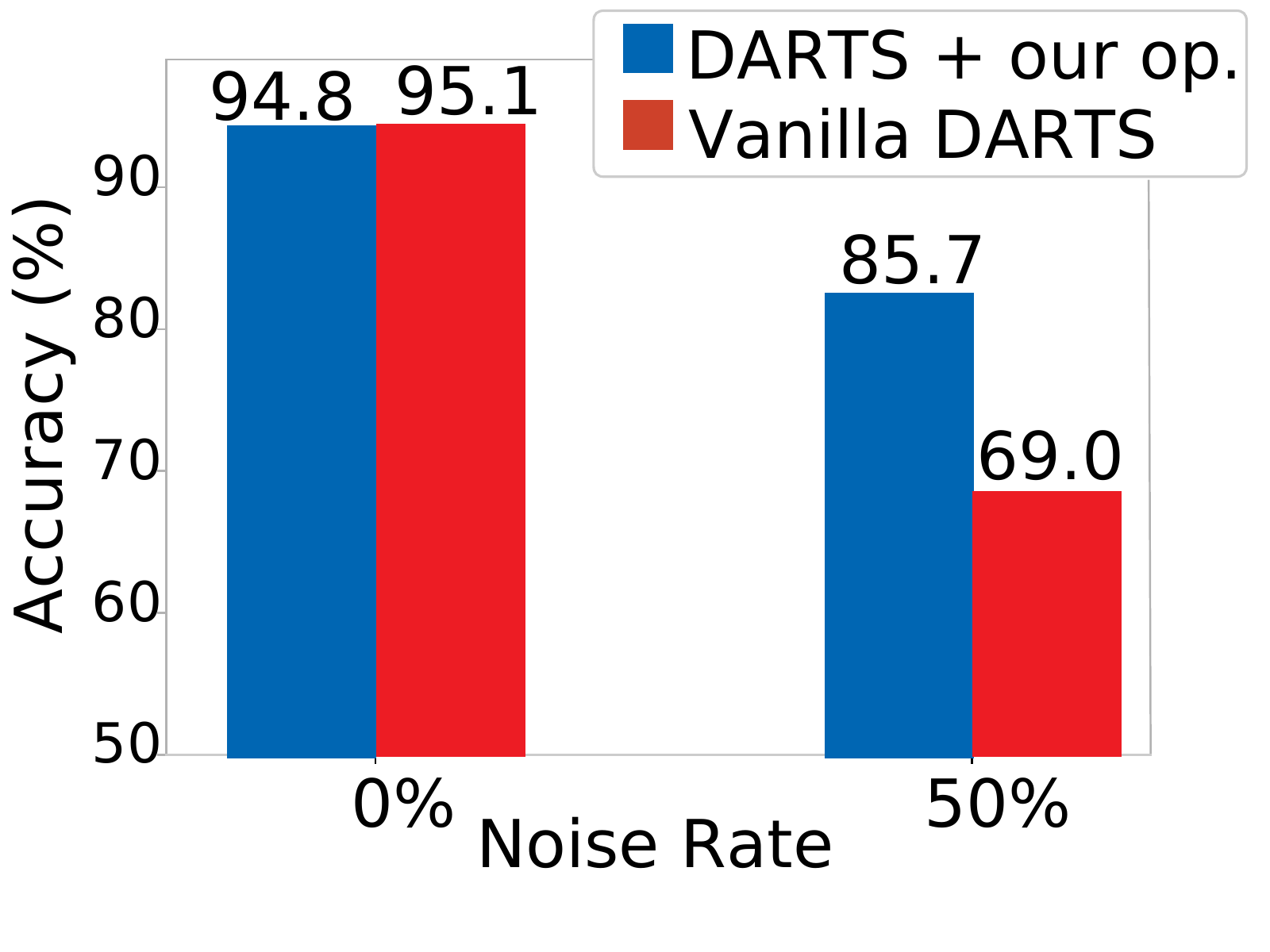}} \quad      
 \subfloat[]{
        \includegraphics[width=0.225\textwidth]{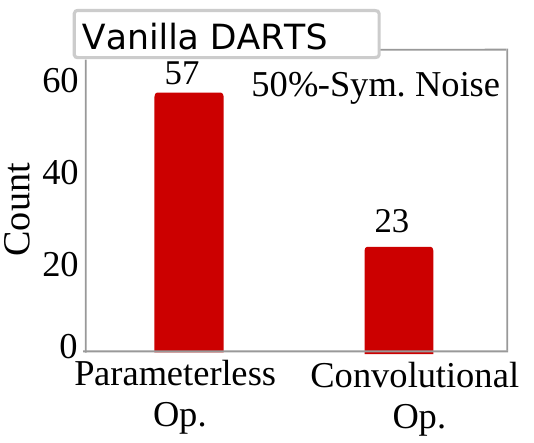}} \quad    
    \subfloat[]{
        \includegraphics[width=0.225\textwidth]{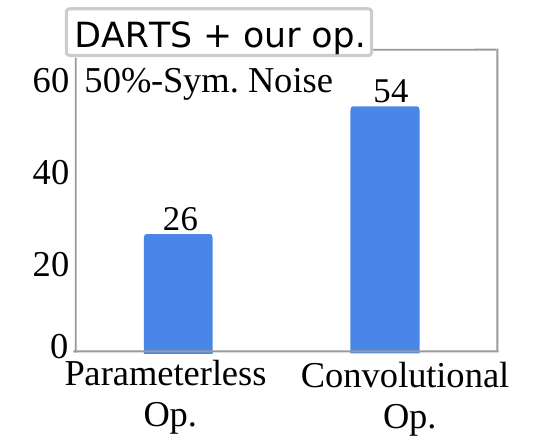}}\quad
    \subfloat[]{
        \includegraphics[width=0.225\textwidth]{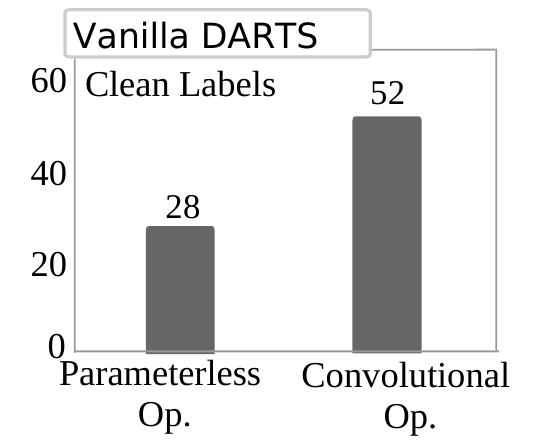}}          
    \caption{(a)  Testing accuracy of vanilla {\em vs.} our approach (CIFAR-10, clean labels {\em vs.} 50\%-symmetric label noise). The histogram of found operators (five runs) of (b) the normal cell of vanilla DARTS under 50\%-symmetric label noise, (c) the normal cell of DARTS with nConv (our noise injecting operator) searched on CIFAR-10 (50\%-symmetric noisy labels), and (d)  the normal cell of vanilla DARTS with clean labels. The normal cell of vanilla DARTS under label noise is constructed poorly due to the large number of parameterless operations. It is apparent the network thus looses the learning capacity in an attempt to prevent overfitting to the noise by selecting parameterless operators. In contrast, DARTS with our proposed operator mitigates such a poor cell design as highlighted by the larger number of convolutional operators being selected in place of a parameterless operator. 
    }
    \label{fig:intro}
\end{figure*}

Our primary focus\footnote{Note that training robust NAS under the label noise is not the same problem as using static network for classification under the label noise.} is  to investigate and advance a robust method to search an architecture in the presence of label noise. To this end, we aim to answer the following questions:
\begin{itemize}[leftmargin=0.6cm]
\item  \textbf{Motivational Curiosity.} As handcrafted neural networks (\eg, Inception~\cite{szegedy2015inception})  overfit to noisy labels~\cite{zhang2016understanding} (even  the pretext labeling in self-supervised UnNAS is imperfect), is the test performance of neural networks constructed by NAS also degraded by the 
label noise?
\item \textbf{Research Curiosity.} Can we design an operator\footnote{An operator is an operation connecting two nodes in NAS \eg, the standard NAS uses conv., max- or average-pooling, skip connections, \etc.} that is robust to noisy labels and helps existing NAS methods (\eg, DARTS~\cite{Liu2018Darts}) to perform well in the presence of label noise?
\end{itemize}

The answer to the first question is affirmative as shown in Fig.~\ref{fig:intro}. 
Firstly, we highlight that the performance of vanilla DARTS~\cite{Liu2018Darts} suffers when subjected to noisy labels as shown in Fig.~\ref{fig:intro} (a). Furthermore, the architecture searched by vanilla DARTS under label noise results in a bad architecture as shown in Fig.~\ref{fig:intro} (b) while our approach shown in Fig.~\ref{fig:intro} (c) produces a better designed cell. To answer the second question, we analyze the task of learning deep neural networks under noisy labels using Information Bottleneck. As a result of this analysis, we introduce a variant of the convolutional operation by injecting noise into the pipeline. Upon learning the parameters of the noise from data, we will empirically show that robustness against label corruption can be attained. In essence, we will show later that the noisy convolution regularizes and limits the gradient of the noisy samples during training.  
In short, our key contributions are: 
\renewcommand{\labelenumi}{\roman{enumi}.}
\hspace{-1.0cm}
\begin{enumerate}[leftmargin=0.6cm]
\item We show that the search through noisy labels degrades the classification performance of standard NAS.
\item We provide an information theoretic framework to tackle learning noisy labels. Our proposed noise injection operator performs implicit regularization during learning preventing overfitting to noisy labels.

\item The proposed operator is included into the NAS search space with the goal of preventing overfitting during training with noisy labels. A noise injection on the input of the operator turns activations of hidden units into the so-called stochastic activations.
\item We show experimentally that architectures emerging from the NAS search with our proposed noise injecting operator outperform under fewer parameters the state of the art, especially in the case of no prior knowledge given \eg, the lack of the noise type/its rates.
\end{enumerate}
\vspace{-0.3cm}

\section{Related work}
\label{sec:related}

\noindent \textbf{Neural architecture search.} 
NAS is the process of automating the design of a neural network architecture. NAS has been successfully applied to various recognition and image generation problems~\cite{Zoph2017NASRL,chen2019detnas,gong2019autogan}.
NAS algorithms can be computationally demanding \eg, when based on evolutionary algorithms~\cite{real2019regularized,elsken2018efficient,esteban2017EvolLarge,stanley2019neuroevol} or reinforcement learning~\cite{Zoph2017NASRL,Baker2016DesignNNRL}, which provide flexible schemes to sample and evaluate architectures from a vast pool of architectures (so-called search space). However, searching directly on a very large dataset with a large neural network is computationally expensive. Zoph \etal~\cite{zoph2018learningscalable} propose a scalable solution by searching a cell structure instead of the entire neural network structure. Due to the above issues, our method follows the formulation of differentiable NAS~\cite{Liu2018Darts,Zela2020RobustDarts} which is computationally feasible on a broad range of tasks~\cite{Lian2020Towards,liu2019auto,Xu2020PC-DARTS} and results in a cell transferrable to other tasks. 

\noindent \textbf{Noisy labels.} 
Label noise can harm the performance of neural networks if training is carried out as if data is clean. One of the solutions to tackle label noise is to relabel the samples as proposed in~\cite{vahdat2017toward,tanaka2018joint,yi2019probabilistic}. Loss correction approaches~\cite{arazo2019unsupervised,reed2015training,jiang2018mentornet,zhang2018generalized} achieve the same purpose of addressing label noise by compensating and modifying the loss functions. Several types of noise (\eg symmetric) can even be modeled as a transition matrix~\cite{patrini2017making,xia2019anchor} that indicates the probability of clean labels being flipped to noisy labels. Other approaches combat noisy labels by ranking~\cite{guo2018curriculumnet}, re-weighting~\cite{ren2018learning} and selecting~\cite{Han2018Co} samples. 
\noindent \textbf{Information bottleneck.}   
Information bottleneck introduced in~\cite{tishby2000information} provides a trade-off formulation between compression of inputs and prediction of outputs. Maximization of the information bottleneck can be approximated by the variational methods~\cite{alemi2016variationalIB}. The variational information bottleneck uses so-called noisy computations for adversarial robustness and it achieves so-called sufficiency, minimality and invariance criteria described in~\cite{achille2018information}. Furthermore, Schwartz-Ziv and Tishby~\cite{shwartz2017opening} propose a visualization technique operating in the so-called \textit{information plane} to analyze the information bottleneck principle in deep learning. The theories and analyses from these prior works inspire our investigations of  robust NAS learning in the presence of label noise under principles of the information bottleneck.

\noindent \textbf{Training neural networks with noise.}
The effect of injecting noise into neural networks has been studied in various contexts. Noise injection as a form of regularization was shown to improve the generalization capability of the model~\cite{bishop1995training, An1996BackpropGrad, gulcehre2016noisy}. Noh \etal~\cite{noh2017regularizing} propose a noise injection method by applying gradient updates over multiple feed-forwards of (random) noisy samples and the gradient is weighted based on the importance per sample in each iteration. The above methods differ from ours \eg, they act on the gradient to prevent saturation or consider optimized dropout. In contrast, we propose a parametric noise injector which prevents overfitting to the label noise and is inspired by the information bottleneck principle.

\section{Robust Differentiable Architecture Search}
\label{sec:proposedmethod}

A NAS algorithm operates in two stages, namely the \textbf{search phase} and the \textbf{evaluation phase}. In the search phase, NAS searches the space of cell architectures given a set of operations. During the evaluation phase, a neural network is constructed by stacking multiple cells prior to re-training it from scratch to evaluate the obtained architecture. In the noisy setting, NAS faces incorrect (noisy) labels during both the search and the evaluation phases. 

To be more specific, let $\mathcal{D}$ be a dataset with $M$ samples 
$\{(\Vec{x}_i, \Vec{y}_i) \}^M_{i=1}$ where $\Vec{x}_i\in\mathbb{R}^n$ and $\Vec{y}_i\in\{0,1\}^C$. 
We define the noisy data as $\mathcal{\tilde{D}}= \{(\Vec{x}_i, \tilde{\Vec{y}}_i)\}$ where $\tilde{\Vec{y}}_i \in\{0,1\}^C$ is a noisy label. There are two types of label noise that we consider in this work, namely symmetric and asymmetric noise. Symmetric noise is constructed by swapping clean labels so that labels of each class are contaminated in a chosen equal proportion by other classes. Asymmetric noise is generated by replacing a chosen percentage of labels  of one class with labels of another class \eg, cat $\rightarrow$ tiger.
The goal of NAS for noisy labels is to find a robust high-performance architecture that copes well with the label noise during training and testing (prevents overfitting to noisy labels).

\subsection{Vanilla DARTS}
In what follows, we build on the DARTS algorithm~\cite{Liu2018Darts} which is the driving force behind several recent developments~\cite{Lian2020Towards,liu2019auto,Xu2020PC-DARTS} due to its significant reduction in computational load of constructing high-performance architectures compared to non-differentiable NAS. 
Prior to introducing our key contributions, we briefly outline DARTS below. 

To perform the search over architectures, DARTS formulates the search phase as a 
bilevel optimization problem. 
To this end, an inner  and an outer objective function $\mathcal{L}_{\mathrm{trn}}$ and  $\mathcal{L}_{\mathrm{val}}$ are introduced with the goal of finding the architecture parameterization $\Vec{\alpha} \in \mathcal{A}$ that minimizes $\mathcal{L}_{\mathrm{val}}$ and the network weights $\Vec{\theta}$ that minimize $\mathcal{L}_{\mathrm{trn}}$ as follows: 
\begin{align}
\begin{split}
    & \min_{\Vec{\alpha} } \; \mathcal{L}_{\mathrm{val}} (\Vec{\theta}^*(\Vec{\alpha}), \Vec{\alpha})
    \\
    & \text{s.t.} \; \Vec{\theta}^\ast(\Vec{\alpha}) = \argmin_\Vec{\theta} \mathcal{L}_{\mathrm{trn}}(\Vec{\theta},\Vec{\alpha}),
\end{split}    
    \label{eq:bilevel}
\end{align}
where $\Vec{\theta}^*(\Vec{\alpha})$ represents the weights of the associated architecture that minimize the inner loss. Optimizing both losses is achieved by gradient descent. Unfortunately,  optimization in the inner loop is time consuming. 
As such, DARTS applies a simple approximation with a single unrolling step to optimize the inner loss 
$ \nabla_\Vec{\alpha} \mathcal{L}_{\mathrm{val}}(\Vec{\theta}^*(\Vec{\alpha}), \Vec{\alpha})  \approx \nabla_\alpha \mathcal{L}_{\mathrm{val}}(\Vec{\theta}^\prime, \Vec{\alpha})$ where $\Vec{\theta}^\prime = \Vec{\theta} - \eta \nabla_\theta \mathcal{L}_{\mathrm{trn}}(\Vec{\theta},\Vec{\alpha})$. In the bi-level formulation, we encounter two derivatives (related to the network parameters and the architecture parameters, respectively) for which the chain rule is applied 
together with the implicit function theorem~\cite{larsen1996design,bengio2000gradient,Foo2008Efficient} which yields: 
\begin{align}
\begin{split}
\!\!\!\!\!\nabla_\Vec{\alpha} \mathcal{L}_{\mathrm{val}}(\Vec{\theta}^*(\Vec{\alpha}), \Vec{\alpha}) 
 &\approx \nabla_\Vec{\alpha} \mathcal{L}_{\mathrm{val}}(\Vec{\theta}^\prime, \Vec{\alpha}) \\ 
\!\!\!\!\! &- \eta \nabla_\Vec{\theta} \mathcal{L}_{\mathrm{val}}(\Vec{\theta}^\prime, \Vec{\alpha})\nabla^2_{\Vec{\theta},\Vec{\alpha}}\mathcal{L}_{\mathrm{trn}}(\Vec{\theta}, \Vec{\alpha}).
\end{split} 
\label{eq:nabla_wrtalpha}
\end{align}

To reduce the training time and computational complexity of the algorithm, one can disregard the higher-order derivatives in Eq.~\ref{eq:nabla_wrtalpha}. This simplification leads to a speed-accuracy trade-off as observed in several studies (\eg,   ~\cite{Liu2018Darts,Zela2020RobustDarts})

Following~\cite{zoph2018learningscalable,real2019regularized,Liu2018Darts}, we make use of two types of cells to design the network, namely \textbf{I.} the normal cell and \textbf{II.} the reduction cell. Each cell is represented by a DAG and can realize one operation from a complete set of candidate operations. The final architecture of the network is constructed by stacking the obtained cells together. 

This enables us to make the composition of the cells both transferable between datasets and independent of the network depth.  A cell is a Directed Acyclic Graph (DAG) with an ordered sequence of $N$ nodes and is connected to its preceding two cells. 
Let $\mathcal{O}$ be the set of candidate operators (\eg, skip connection, $3\times3$ dilated convolution, \etc) defined for a cell. Consider a node $i$ whose output represented by $\Vec{z}^{(i)}$ is obtained by:
\begin{align}
    \Vec{z}^{(i)} = \sum_{j \leadsto i} o^{(i,j)}\big(\Vec{z}^{(j)}\big)\;,
    \label{eq:hiddenunit1}
\end{align}
where $\leadsto$ denotes  the edge between nodes $i$ and $j$ with a total weight over possible operations given as: 
\begin{align}
o^{(i,j)}(\Vec{z}^{(j)}) = \sum_{r =1}^{\nu} \frac{\exp \Big(\alpha^{(i,j)}_r\Big)}
{\sum_{s = 1}^{\nu} \exp\Big(\alpha^{(i,j)}_{s}\Big)} o^{(i,j)}_r(\Vec{z}^{(j)})\;,
    \label{eq:hiddenunit2}
\end{align}
\noindent
where $o^{(i,j)}_r$ are individual operations. In essence, choosing the operation that will connect node $j$ to node $i$ is formulated by 
a mixing weight vector $\Vec{\alpha}^{(i,j)} = (\alpha^{(i,j)}_1,\alpha^{(i,j)}_2,\cdots,\alpha^{(i,j)}_\nu)$ using a softmax function.
The task of architecture search is then to learn  $\Vec{\alpha} = \big\{ \alpha^{(i,j)} \big\}$. At the end of the search phase, a discrete architecture is picked by choosing  the most likely operation between the nodes. That is, $o^{(i,j)} = o^{(i,j)}_{r^\ast}$ where $r^\ast = \argmax_r \alpha^{(i,j)}_r$.

\begin{figure*}[t]
    \centering
     \subfloat[]{
    \includegraphics[width=0.3\textwidth]{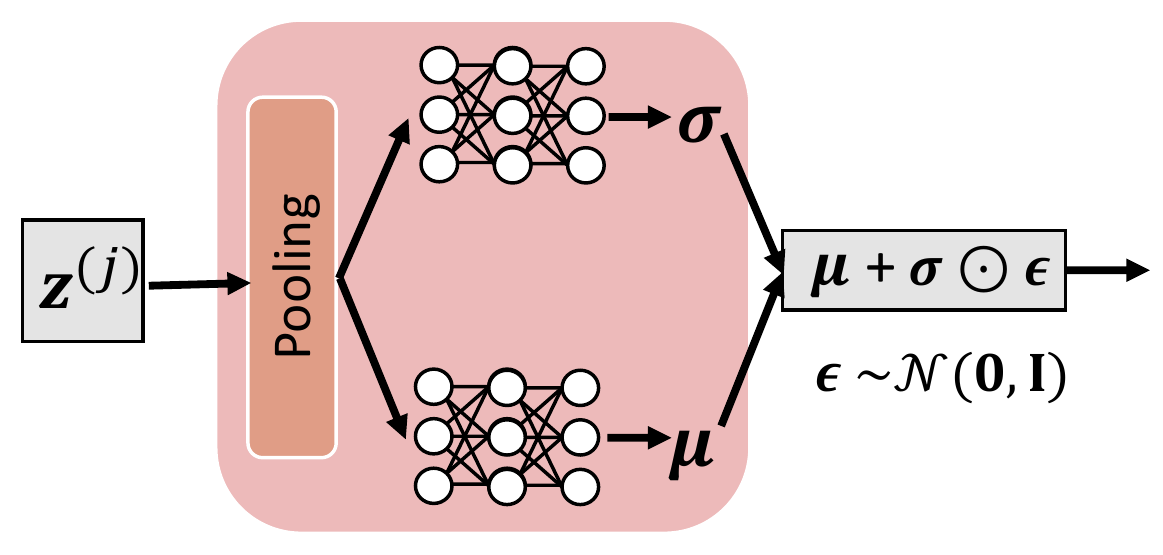}}  \quad \quad
    \subfloat[]{
        \includegraphics[width=0.24\textwidth]{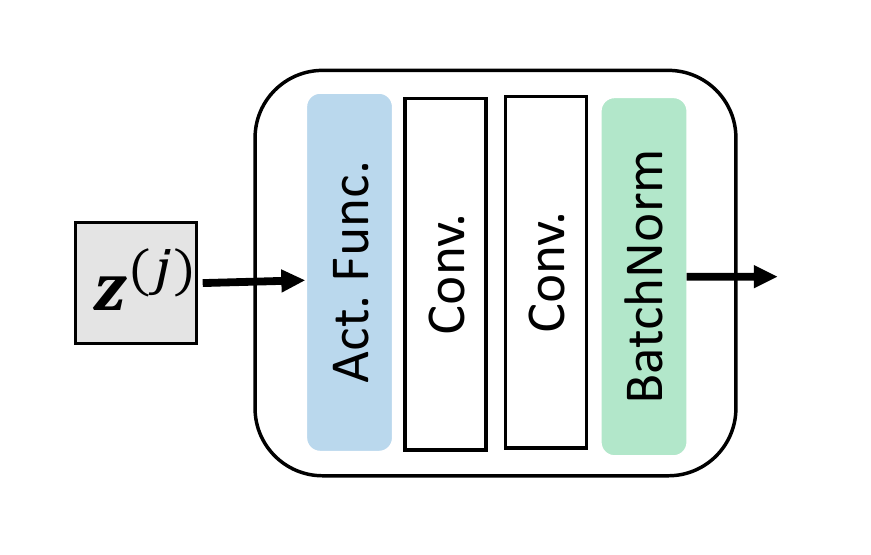}}   
       \quad   \quad
    \subfloat[]{
    \includegraphics[width=0.24\textwidth]{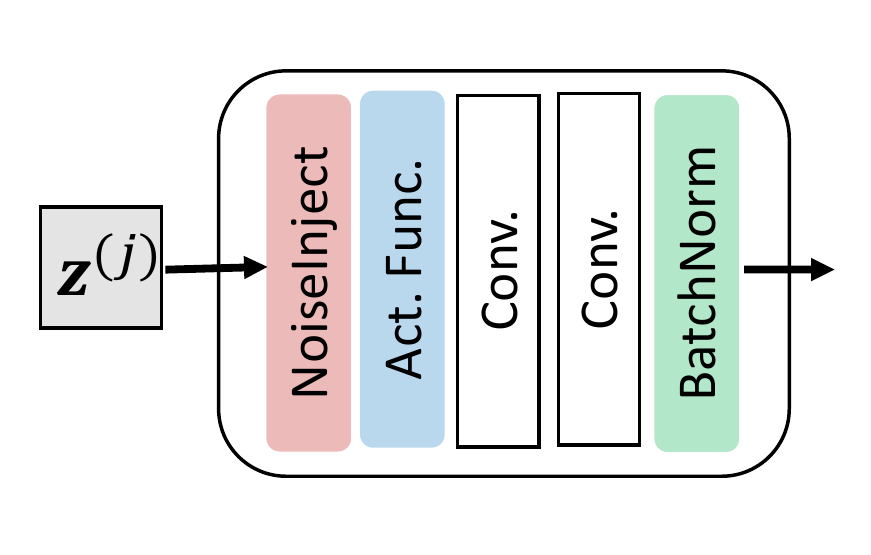}}    
    \caption{ (a) The noise injection module which is incorporated into a convolutional operator. (b) The basic convolutional operator used in DARTS (c) Our  convolutional operator with additional noise injection at the input $\Vec{h}$. 
    }\vspace{-0.35cm}
    \label{fig:operator}
\end{figure*}

\subsection{Learning in the presence of label noise via the information bottleneck principle}
Differentiable NAS uses the cross-entropy loss to train (search and evaluation) networks. Below, we formulate a loss function which prevents the trained model from overfitting to noisy labels. To this end, we bring the information theoretic view and explain how it helps with training under label noise. We begin by defining several terms in our analysis: the input $\Vec{x} \in X$, the output $\Vec{y} \in Y$, the representation of the input $\Vec{z} \in Z$, entropy $H(X)=\mathbb{E}_{p(\Vec{x})}[-\log p(\Vec{x})]$, cross-entropy $H({p, q})=\mathbb{E}_{p(\Vec{x})}[-\log q(\Vec{x})]$, conditional entropy $H(Y \vert X)=H(X,Y)-H(X)$, mutual information $I(X;Y)=H(Y) - H(Y \vert X)$, and Kullback-Leibler (KL) divergence ${\text{KL}(p(\Vec{x})\|q(\Vec{x}))=\mathbb{E}_{p(\Vec{x})}[\log \frac{p(\Vec{x})}{q(\Vec{x})}]}$. 

Considering our NAS setting, we 
revisit the Information Bottleneck~\cite{tishby2000information} principle which provides a computational framework that controls the trade-off between the compression of input $\Vec{x}$ and the prediction of $\Vec{y}$ according to:
\begin{equation}
\max_{\Vec{\phi}} \;I(Z;Y, \Vec{\phi}) - \beta I(X;Z, \Vec{\phi}),   
\label{eq:info_bottleneck}
\end{equation}
where $\beta \geq 0$ is a hyper-parameter adjusting the level of relevant information captured by $Z$ while $\Vec{\phi}$ denotes network parameters. For training neural networks under the regular setting (clean labels), one maximizes $I(Z;Y, \Vec{\phi})$ to obtain a model with a high performance. However, increasing the value of $I(Z;Y, \Vec{\phi})$ alone makes the model prone to overfitting~\cite{achille2018emergence} \ie, the network memorizes the dataset and overfits to wrongly labeled datapoints which results in a performance degradation, as shown in our experiments \textsection~\ref{sec:experiments}. 
Let us consider the clean and noisy labels $\BLUE{{\hat{Y}}}$ and  $\RED{{\widetilde{Y}}}$ separately by splitting the first term in~Eq.~\ref{eq:info_bottleneck} which leads to:
%
%
%
\begin{eqnarray}
\!\!\!\!\!\!\!I(Z;Y, \Vec{\phi})&\!\!\!=\!\!\!&  I(Z;\BLUE{\hat{Y}}, \Vec{\phi}) +  I(Z;{\RED{\widetilde{Y}}}, \Vec{\phi}) \nonumber\\
\!\!\!\!\!\!\!&\!\!\!=\!\!\!& \!\!\int p(\Vec{\BLUE{\hat{y}} ,\Vec{z} \vert  \Vec{\phi}} ) \log \frac{p(\Vec{\BLUE{\hat{y}}},\Vec{z}\vert  \Vec{\phi})}{p(\Vec{\BLUE{\hat{y}}}\vert  \Vec{\phi})p(\Vec{z}\vert  \Vec{\phi})} \;d\Vec{\BLUE{\hat{y}}}\;d\Vec{z} \nonumber\\
\!\!\!\!\!\!\!&& \!\!\!\!\!\!\!\!+ \int p(\Vec{\RED{\widetilde{y}}},\Vec{z}\vert  \Vec{\phi}) \log \frac{p(\Vec{\RED{\widetilde{y}}},\Vec{z}\vert  \Vec{\phi})}{p(\Vec{\RED{\widetilde{y}}}\vert  \Vec{\phi})p(\Vec{z}\vert  \Vec{\phi})}\;d\Vec{\RED{\widetilde{y}}}\;d\Vec{z}.\label{eq:simple}
\end{eqnarray}
Substituting $ p(\Vec{y},\Vec{z}) = {p(\Vec{y} \vert \Vec{z})}{p(\Vec{z})}$ in Eq. \ref{eq:simple}, we obtain:
\begin{eqnarray}
\!\!\!\!    I(Z;\BLUE{\hat{Y}}, \Vec{\phi})  &=& \int p(\Vec{\BLUE{\hat{y}}},\Vec{z}\vert \Vec{\phi}) \log \frac{p(\Vec{\BLUE{\hat{y}}} , \Vec{z}\vert \Vec{\phi} )}{p(\Vec{\BLUE{\hat{y}}}\vert \Vec{\phi})} \;d\Vec{\BLUE{\hat{y}}}\;d\Vec{z} \nonumber\\
\!\!\!\!    I(Z;{\RED{\widetilde{Y}}}, \Vec{\phi}) &=&  \int p(\Vec{\RED{\widetilde{y}}}, \Vec{z}\vert \Vec{\phi}) \log \frac{p(\Vec{\RED{\widetilde{y}}} \vert \Vec{z}; \Vec{\phi})}{p(\Vec{\RED{\widetilde{y}}}\vert \Vec{\phi}) }\;d\Vec{\RED{\widetilde{y}}}\;d\Vec{z} . 
\end{eqnarray}
As estimating terms ${p(\Vec{\RED{\widetilde{y}}} \vert \Vec{z}; \Vec{\phi})}$ and ${p(\Vec{\BLUE{\hat{y}}} \vert \Vec{z}; \Vec{\phi})}$ of the mutual information is often computationally intractable, it can be completed with the variational approximation. 
We note that:
\begin{eqnarray}
          I(Z;\BLUE{\hat{Y}}, \Vec{\phi}) &\geq& \int p(\Vec{\BLUE{\hat{y}}},\Vec{z} \vert \Vec{\phi}) \log \frac{\bar{p}(\Vec{\BLUE{\hat{y}}} \vert \Vec{z}; \Vec{\phi})}{p(\Vec{\BLUE{\hat{y}}}\vert \Vec{\phi})} \;d\Vec{\BLUE{\hat{y}}}\;d\Vec{z}, \nonumber\\
    I(Z;{\RED{\widetilde{Y}}}, \Vec{\phi}) &\geq&  \int p(\Vec{\RED{\widetilde{y}}}, \Vec{z}\vert \Vec{\phi}) \log \frac{\bar{p}(\Vec{\RED{\widetilde{y}}} \vert \Vec{z}; \Vec{\phi})}{p(\Vec{\RED{\widetilde{y}}}\vert \Vec{\phi}) }\;d\Vec{\RED{\widetilde{y}}}\;d\Vec{z}. \nonumber
\end{eqnarray}
 We formulate the lower bound for the information bottleneck with noisy labels as follows:
\begin{align}
\begin{split}
I(Z;\BLUE{\hat{Y}}, \Vec{\phi}) +  I(Z;{\RED{\widetilde{Y}}}, \Vec{\phi}) - \beta I(X;Z, \Vec{\phi}) \geq \\
\int p(\Vec{x})p(\BLUE{\Vec{\hat{y}}} \vert \Vec{x}; \Vec{\phi})p(\Vec{z} \vert \Vec{x}; \Vec{\phi}) \log \bar{p}(\BLUE{\Vec{\hat{y}}} \vert \Vec{z}; \Vec{\phi}) \;d\Vec{x} \;d\Vec{\hat{y}}\; d\Vec{z}  \\
+ \int p(\Vec{x})p(\RED{\Vec{\widetilde{y}}} \vert \Vec{x}; \Vec{\phi})p(\Vec{z} \vert \Vec{x}; \Vec{\phi})
\log \bar{p}(\RED{\Vec{\widetilde{y}}} \vert \Vec{z}; \Vec{\phi}) \;d\Vec{x} \;d\Vec{\widetilde{y}}\; d\Vec{z} \\
- \beta\int p(\Vec{x})p(\Vec{z}|\Vec{x}; \Vec{\phi})\mathrm{log}\frac{p(\Vec{z} \vert \Vec{x}; \Vec{\phi})}{p(\Vec{z}\vert \Vec{\phi})} \;d\Vec{x}\; d\Vec{z}
= \mathcal{L}. 
\end{split}
\end{align}
Let $q(\Vec{z} \vert \Vec{x}; \Vec{\phi})$ and $r(\Vec{z})$ be a variational approximation for $p(\Vec{z} \vert \Vec{x}; \Vec{\phi})$ and $p(\Vec{z} \vert \Vec{\phi})$, then we have the lower bound:
\begin{align}
\begin{split}
\mathcal{L} \approx \frac{1}{|\BLUE{\hat{Y}}| + |\RED{\widetilde{Y}}|} \sum_{n=1}^{|\BLUE{\hat{Y}}| + |\RED{\widetilde{Y}}|} \mathbb{E}_{\Vec{z} \sim q(\Vec{z} \vert \Vec{x}; \Vec{\phi})}\big{[} \log p(\Vec{y}_n | \Vec{z}) \big{]} \\
- \beta \text{KL}\big{[} q(\Vec{z}|\Vec{x}_n; \Vec{\phi}) \vert \vert r(\Vec{z}) \big{]},
\end{split}
\label{eq:variational_IB}
\end{align}
 where $q(\Vec{z}|\Vec{x};\Vec{\phi})$ denotes our variational approximation using the reparameterization trick~\cite{kingma2013auto,alemi2016variationalIB}. 
 Given the case with noisy labels, we can reduce $I(Z;\RED{\widetilde{Y}}, \Vec{\phi})$ by controlling the first term in~Eq.~\ref{eq:variational_IB} to avoid overfitting in the training phase. In fact, we can control 
 the first term by adjusting the noise variance of the second term of~Eq.~\ref{eq:variational_IB} as follows.

Suppose an encoder is used for $q(\Vec{z}|\Vec{x}; \Vec{\phi})$, $r(\Vec{z}) = \mathcal{N}(\Vec{0}, \textbf{I})$  and $\Vec{\phi}=\{\Vec{\phi}_1,\Vec{\phi}_2\}$. We can calculate the KL divergence in the closed form as follows: 
\begin{align}
        \text{KL}(q(\Vec{z}|\Vec{x}; \Vec{\phi}) \vert \vert r(\Vec{z})) &= - \sum_j^J \log {{\sigma_j}^2} + \frac{{\sigma_j}^2 + {\mu_j}^2}{2} - \frac{1}{2},
\end{align}
where $\Vec{z} \sim \mathcal{N}(\Vec{\mu}_\Vec{x}, \Vec{\sigma}^2_\Vec{x}) = q(\Vec{z}|\Vec{x}; \Vec{\phi})$, $\Vec{\mu}_\Vec{x} = f_\Vec{\phi_1}(\Vec{x})$, $\Vec{\sigma}_\Vec{x} = f_\Vec{\phi_2}(\Vec{x})$, and $j$ is the index element of  $\Vec{\mu}_\Vec{x}$ and $\Vec{\sigma}_\Vec{x}$.

\subsection{DARTS meets the label noise}

The variational information bottleneck  provides an inspiration for our noise injecting operators for differentiable NAS. It helps us limit overfitting and memorization to noisy labels. 
We argue that the designed operator should exhibit the variational approximation in~Eq.~\ref{eq:variational_IB} to adjust $I(Z;\RED{\widetilde{Y}})$. To this end, a robust operator should inject the right levels of noise to successfully train NAS in the presence of label noise. Thus, we propose a new operator, the noise injecting Convolution, \emph{nConv} for short. In NAS, the architecture parameters $\Vec{\alpha}$ play an important role in assigning weights of operators. As we train the model under label noise, the larger the weights corresponding to our noise injecting operator are, the stronger the regularization effect becomes. 

 We define the functionality of our nConv operator on its input $\Vec{z}$ as 
$\Vec{z} \sim \mathcal{N}(\Vec{\mu}_\Vec{x}, \Vec{\sigma}^2_\Vec{x})$\footnote{We have modeled the Normal distribution by a diagonal covariance. However, our algorithm is generic and can deal with full covariance matrices. We denote the diagonal elements of the covariance matrix by $\Vec{\sigma^2}$. 
}. By properly scaling the noise parameters, we expect a neural network with nConv  to exhibit robustness against noisy labels. To this end, we propose  to learn the noise parameters (\ie, $\Vec{\mu}_\Vec{x}$ and $\Vec{\sigma}^2_{\Vec{x}}$) based on the data distribution. 
Drawing inspiration from the reparameterization trick~\cite{kingma2013auto}, 
this is achieved by learning $\Vec{\mu}$ and $\Vec{\sigma^2}$ through a network parameterized by $\Vec{\phi}$ as $\Vec{z} \sim q_(\Vec{z} \vert  \Vec{x}; \Vec{\phi})$ (see Fig.~\ref{fig:operator}~(a) for a conceptual diagram). In DARTS, $\Vec{\phi}$ corresponds to noise parameters of all instantiated noise injecting operators. Let us assume $\Vec{\phi}$ is contained within $\Vec{\theta}$ represents the entire learnable set of parameters of all instantiated operators. As in Eq.~\ref{eq:variational_IB}, the problem of learning the parameters of noise is cast as 
marginalizing the negative log-likelihood over the data with a KL divergence regularization term as:
\vspace{-0.15cm}
\begin{align}
\!\!
\begin{split}
\mathcal{L}_{\mathrm{NAS}}=
\hspace{-1ex}\sum_{(\Vec{x},\tilde{\Vec{y}}) \in \tilde{D}} \hspace{-1ex}-\mathbb{E}_{\Vec{z} \sim q(\Vec{z} \vert \Vec{x}; \Vec{\alpha}; \Vec{\theta} )} 
 \big[\log  p\big{(}\tilde{\Vec{y}} \vert  \Vec{x}; \Vec{\theta}; \Vec{\alpha}; \Vec{z}\big{)} \big] \\
 + \;\beta \text{KL}\big[q\big(\Vec{z} \vert \Vec{x}; \Vec{\alpha}; \Vec{\theta} \big) \vert \vert \mathcal{N}\big(\Vec{0}, \mathbf{I}\big)\big]\;.
\end{split}\!\!
\label{eq:minimize}
\vspace{-0.15cm}
\end{align}
Note that, minimizing Eq.~\ref{eq:minimize} also applies when we retrain the network in the evaluation phase by using a discrete version of $\Vec{\alpha}$. Algorithm~\ref{code:algorithm} details the steps of performing DARTS inclusive of our nConv operator.
\begin{algorithm}
\caption{DARTS + nConv}
Operation $\mathcal{O}$, arch. parameters $\Vec{\alpha}$, network parameters $\Vec{\theta}$, noise injection parameters $\Vec{\phi}$, dataset $\mathcal{\tilde{D}}$  

\begin{algorithmic}[1]
\State \textbf{Phase 1: Search an architecture}
    \While{not converged}
    \State Update $\Vec{\alpha}$ using Equation~\ref{eq:nabla_wrtalpha}
    \State Update $\Vec{\theta}, \Vec{\phi}$ using $\nabla_{\Vec{\theta}, \Vec{\phi}} \mathcal{L}_{\mathrm{NAS}}$
    \EndWhile
    \State \textbf{Phase 2: Evaluate an architecture}
    \State Construct a final architecture from $\Vec{\alpha}$
    \State Reinitialize $\Vec{\theta}, \Vec{\phi}$
    \While{not converged}
    \State Update $\Vec{\theta}, \Vec{\phi}$ using $\nabla_{\Vec{\theta}, \Vec{\phi}} \mathcal{L}_{\mathrm{NAS}}$
    \EndWhile
     
\end{algorithmic}

\label{code:algorithm}
\end{algorithm}

\begin{figure*}[h]
    \centering
 \subfloat{
        \includegraphics[width=0.23\textwidth]{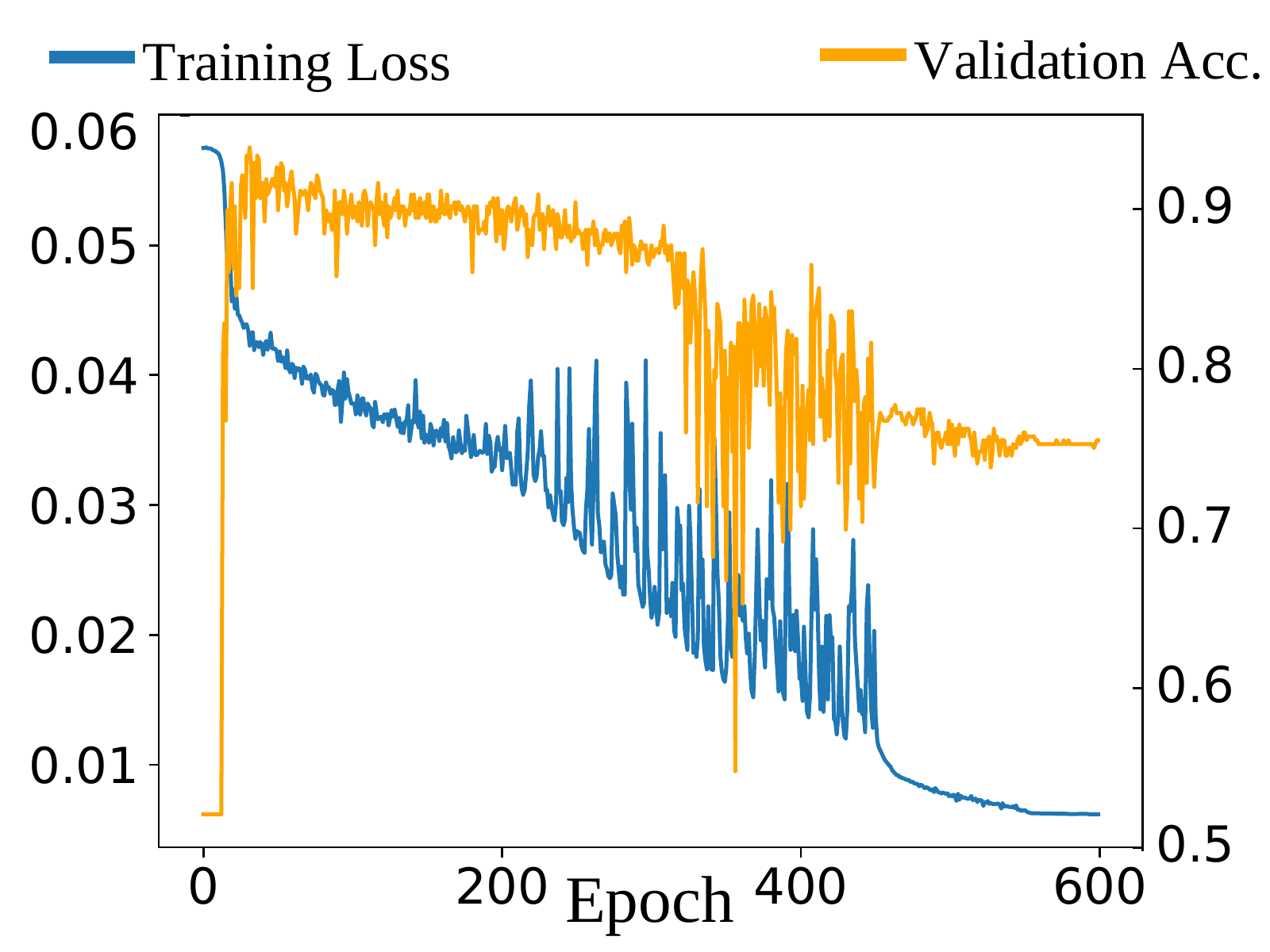}} \quad    
 \subfloat{
        \includegraphics[width=0.22\textwidth]{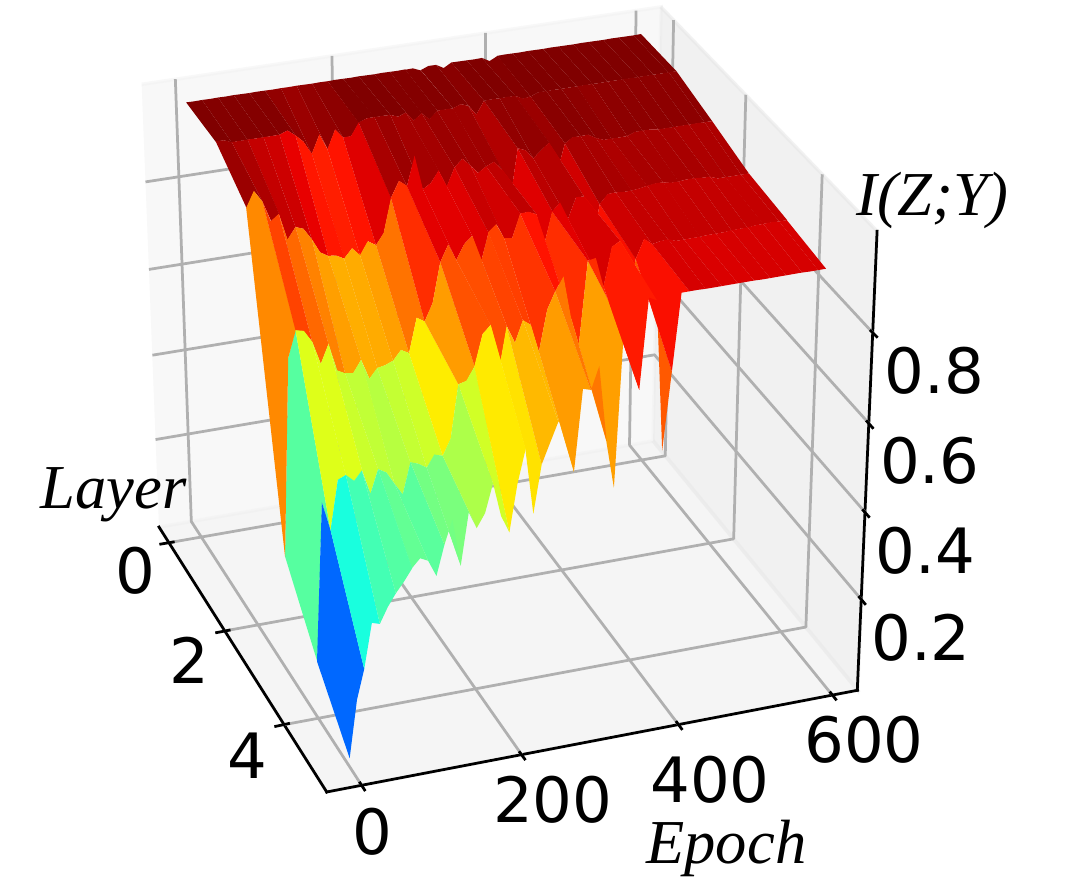}}
        \quad
 \subfloat{
        \includegraphics[width=0.22\textwidth]{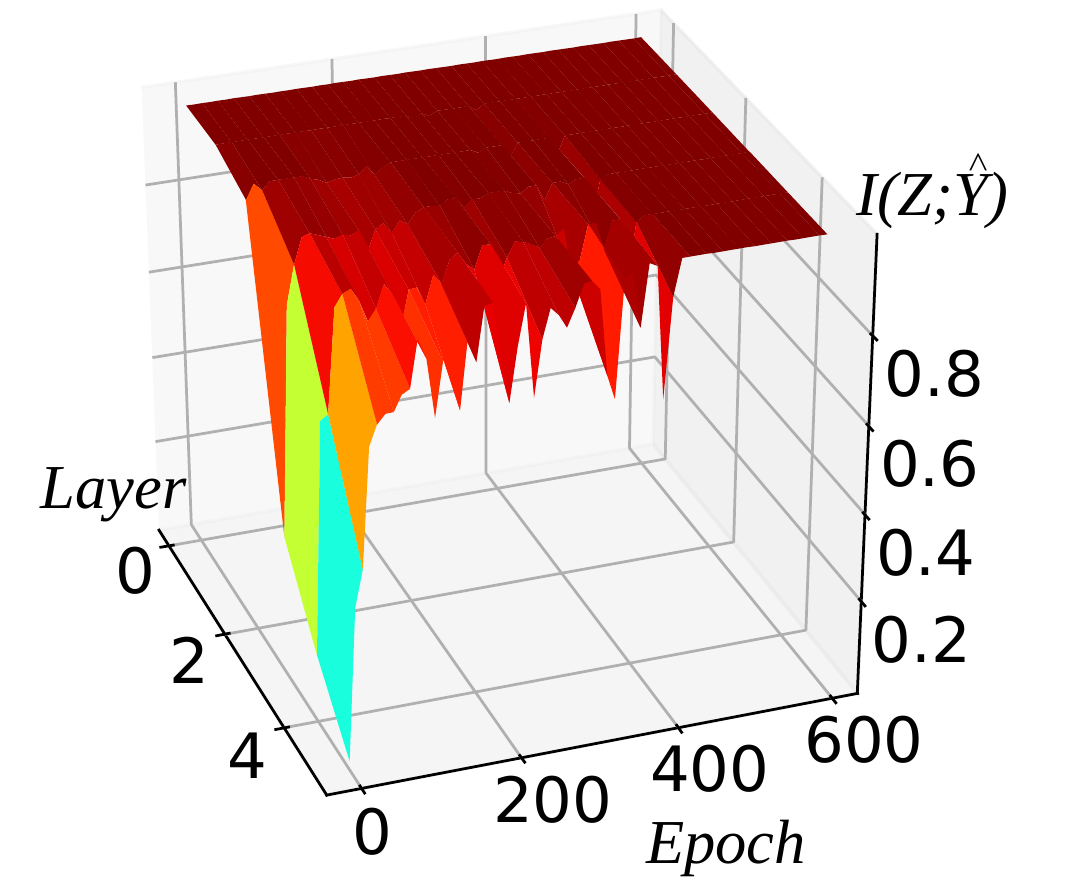}}       \quad
 \subfloat{
        \includegraphics[width=0.245\textwidth]{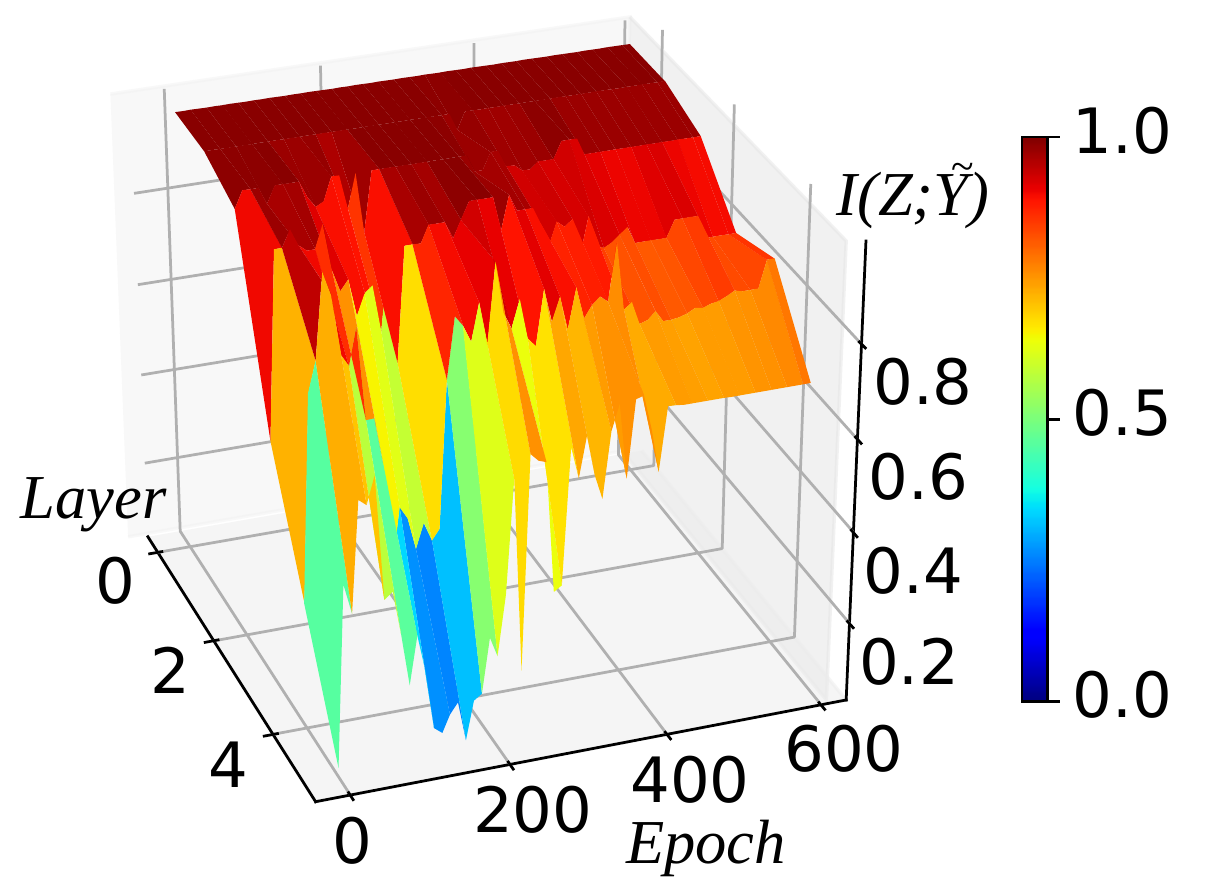}}
        \\ \setcounter{subfigure}{0}
 \subfloat[]{\includegraphics[width=0.23\textwidth]{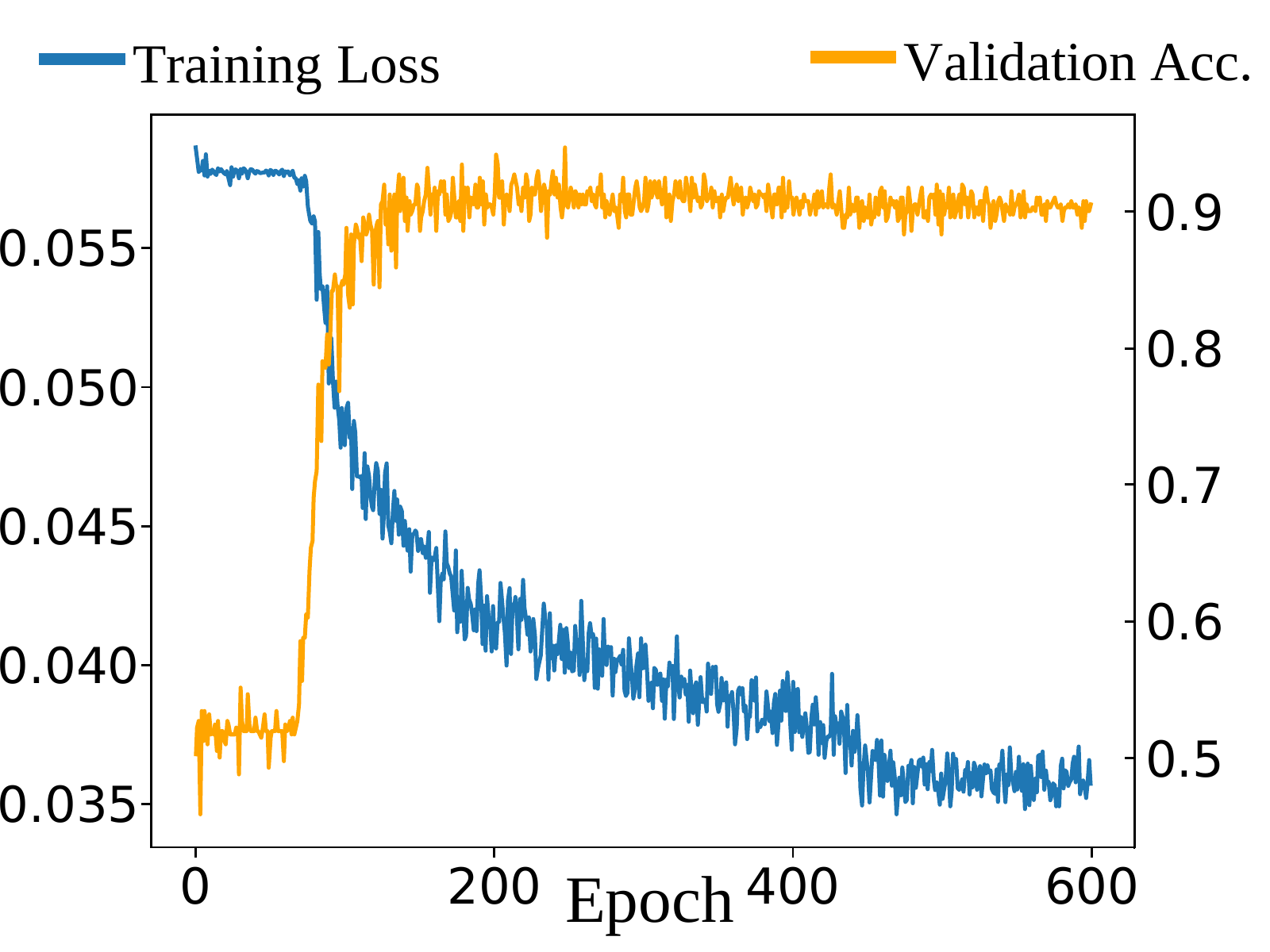}} \quad 
 \subfloat[]{\includegraphics[width=0.22\textwidth]{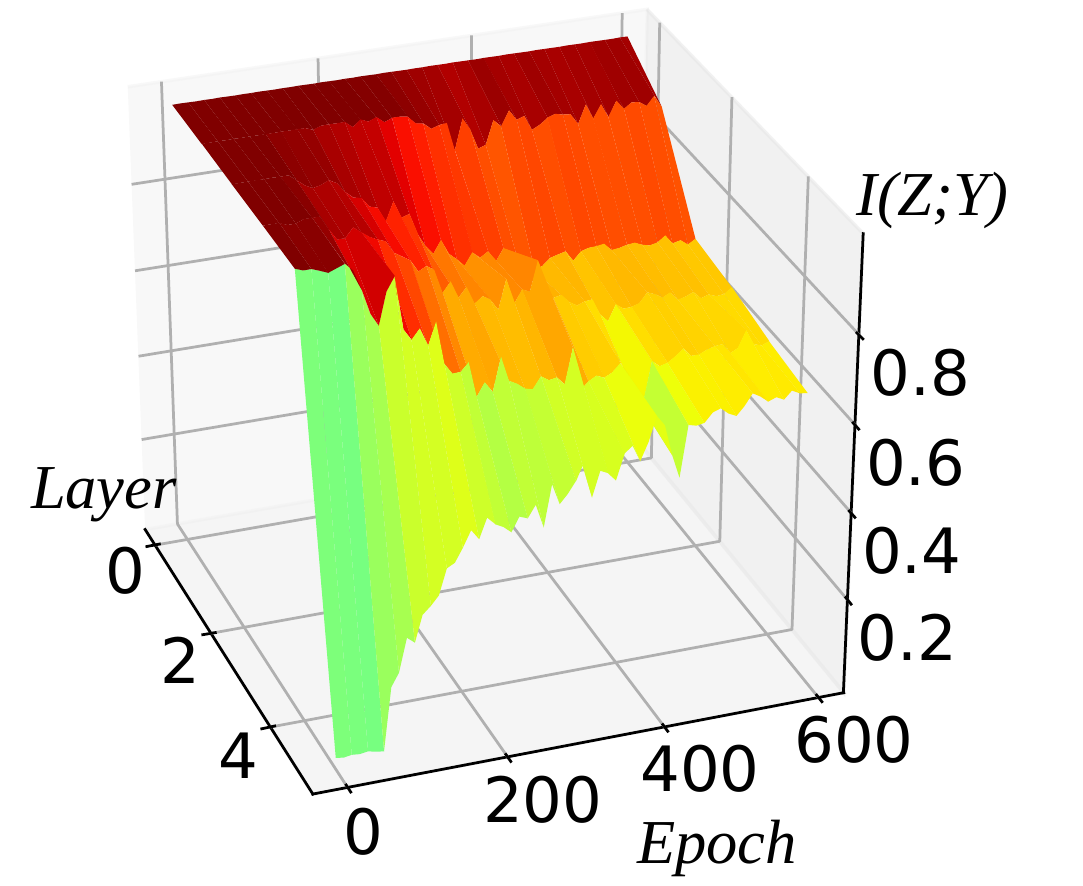}} 
        \quad 
 \subfloat[]{\includegraphics[width=0.22\textwidth]{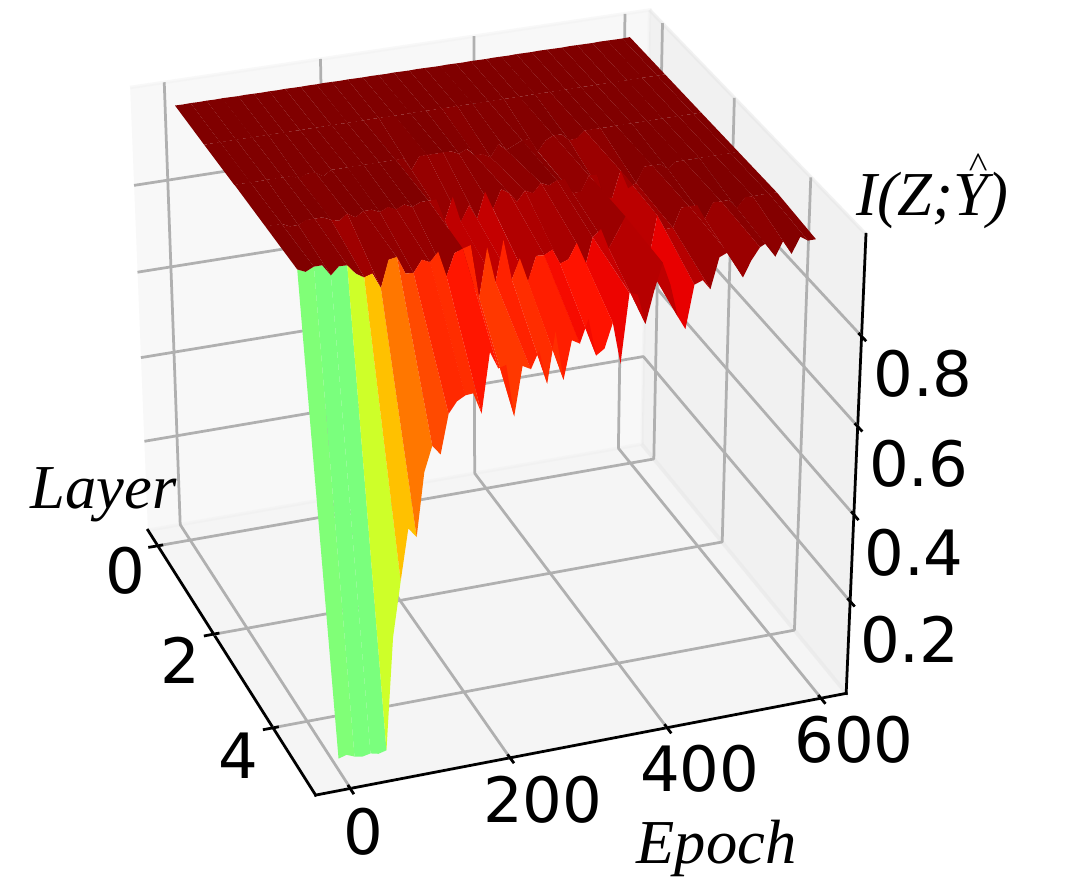}}
        \quad 
\subfloat[]{
\includegraphics[width=0.245\textwidth]{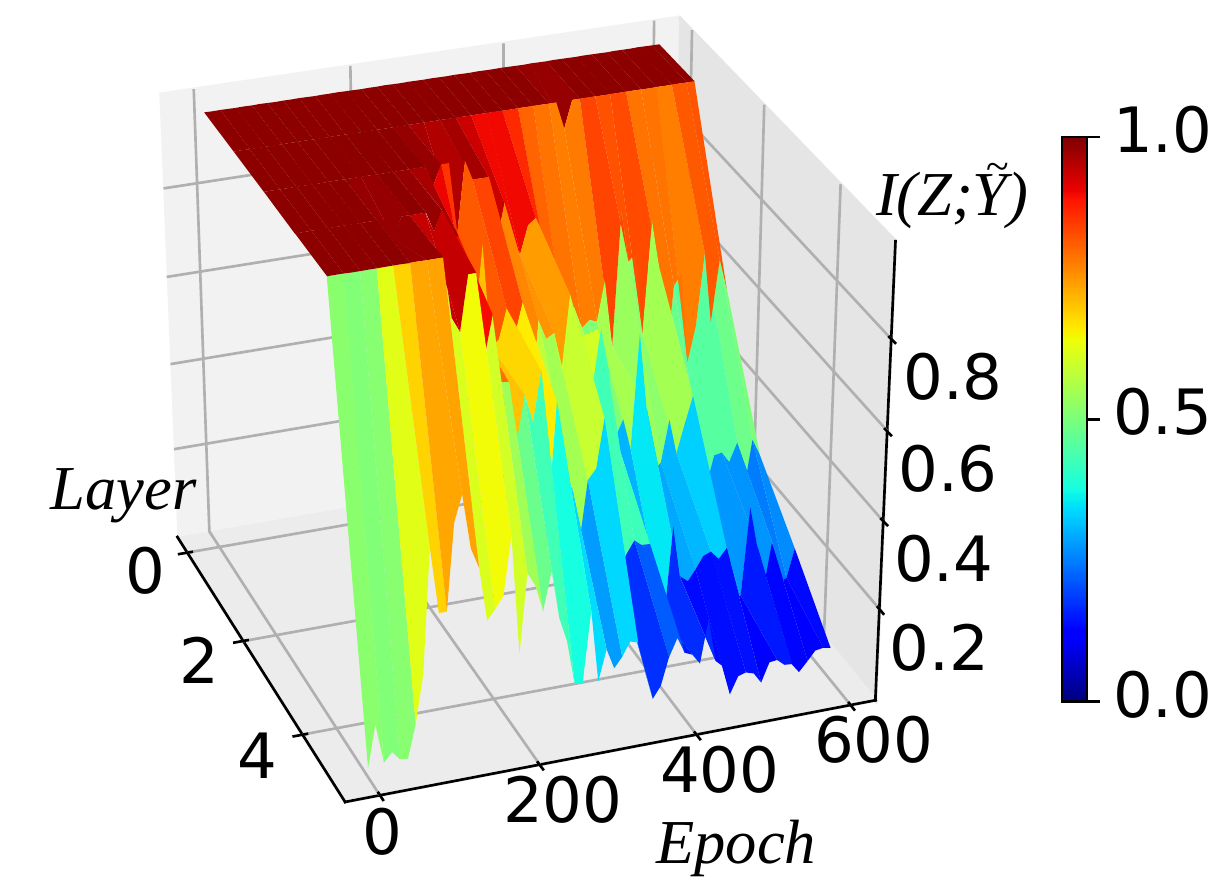}}   
    \caption{Visualization of  mutual information for noisy labels. (Top) Training under 20\% label noise without (top) and with our approach (bottom). (a) Training loss and validation accuracy across 600 epochs. (b) Mutual information of all labels and the hidden units ($I(Z;Y)$). (c) Mutual information of clean labels with hidden units. (d) Mutual information of noisy labels with hidden units. The higher the value of $I(Z;Y)$, the more information about $Y$ is preserved by variable $Z$.
    }
    \label{fig:visualization_mi}
\end{figure*}
In vanilla DARTS, two blocks of convolution, namely \emph{Separable Convolution} (SepConv) and \emph{Dilated Convolution} (DilConv) are used to construct the search space. 
Both blocks share the same architectural design (see Fig.~\ref{fig:operator}~(b)). The variants of these two blocks are constructed by adjusting the kernel size, the dilation size, and the stride size of the underlying convolutions. We extend this basic block by injecting noise at the input $\Vec{z}^{(j)}$ (see Fig.~\ref{fig:operator}~(c)). We can readily extend both blocks with nConv to attain \emph{SepNConv} and \emph{DilNConv} blocks. At the cell level, both normal and reduction cells may include nConv in their structures. In the normal cell, the feature map is generated with the same dimension as the input. For the reduction cell, the input to the noise injection module is applied with a stride of two to shrink the cell output size by a factor of two. Selecting nConv in the cells encourages stochastic activations (the output of hidden units) that help regularize the neural network. Note that we experimentally observe that multiplication of noise and hidden units (\eg dropout) in nConv does not work well to tackle the label noise problem. Thus, our noise injection function is defined by addition of noise to the hidden units. 

\begin{tcolorbox}[width=1.0\linewidth, colframe=blackish, colback=beaublue, boxsep=0mm, arc=3mm, left=1mm, right=1mm, right=1mm, top=1mm, bottom=1mm]
\textbf{Theoretical Interpretation.} 
We wire up a network under noise conditions. To this end, it does not suffice to merely tackle the label noise at the classifier layer. From the information bottleneck, the model with the noise injector emerges $\mathbf{x}_{l+1}\!=\!f(\mathbf{x}_l;\boldsymbol{\Theta}^\mu_l)\!+\!\mathcal{N}(\Vec{0}, \mathbf{I})g(\mathbf{x}_l;\boldsymbol{\Theta}^\sigma_l)$.
In effect, $g(\cdot)$ can be understood as modeling the variance by $\boldsymbol{\Theta}^\sigma_l$ and hence, the sum  $\mathbf{x}_{l+1}\!=\!f(\cdot)+\mathcal{N}(\Vec{0}, \mathbf{I})g(\cdot)$  pushes $\mathbf{x}_{l+1}$ to contain the information about the parameter uncertainty.
The uncertainty flows through the layers
as shown for example in Fig.~\ref{fig:operator}. This propagation of uncertainty prevents trivial skip-connections. 
\end{tcolorbox}

\section{Experiments}
\label{sec:experiments}

\subsection{The dynamics of training under label noise}
In this experiment, we estimate $I(Z;Y)$ to analyze the mutual information between the hidden units of neural networks (with and w/o noise injecting operators) and their respective labels. The settings are adopted from~\cite{shwartz2017opening} which provides a toy dataset and a technique to visualize the mutual information. The network for this experiment has six layers in total with Tanh activations on intermediate layers and a Sigmoid function for final outputs. We contaminate 20\% training data from the toy dataset~\cite{shwartz2017opening} that has only two categories. Fig.~\ref{fig:visualization_mi} shows that training the network with our approach ($\beta=1$) under label noise is more stable across 600 epochs. The mutual information is indicative of a correlation between two variables. Fig.~\ref{fig:visualization_mi}~(d) shows that a neural network w/o noise injecting operator overfits to noisy labels (top) while our proposed noise injection reduces overfitting (bottom). This experiment confirms our theoretical analysis that the variational information bottleneck, in essence, reduces the mutual information between the hidden units and noisy labels to combat memorization of corrupted labels. 

\subsection{Searching the architecture} 
We search an architecture with the following operators $\mathcal{O}$: $3\times3$ separable convolution, $3\times3$ dilated convolution,  $3\times3$ separable convolution with noise, $3\times3$ dilated convolution with noise, $3 \times 3$ max pooling, $3 \times 3$ average pooling, and the identity. Separable and dilated convolution follows the operations in~\cite{zoph2018learningscalable,real2019regularized,Liu2018Darts}. The architecture search is run on CIFAR-10 consisting of training and validation sets. The setup is similar to DARTS in that there are 7 nodes for each cell and 16 initial filters. We use the momentum SGD, we set the learning rate and the momentum to 0.025 and 0.9, resp., and the weight decay to 3$\times$ $10^{-4}$. Moreover, a cosine schedule is  employed to anneal the learning rate. A network with 8 cells is trained for 50 epochs with a batch size of 96. The search time is 6 hours on a NVIDIA Titan V GPU.

\subsection{Datasets and implementations details}
The evaluation phase follows the protocol in DARTS in which the architecture is found via a small task and then the cell structure is transferred to other (bigger) tasks. Normal cells and reduction cells are based on the architecture found on CIFAR-10, then the cells are stacked for evaluation. All experiments for the architecture evaluation are averaged over 3 runs. The model is trained for 300 epochs for all datasets. In the evaluation phase, the cells are stacked to form a group of 16 cells and 16 initial filters. The optimizer parameters are set following the setup during the search phase.  

Benchmark datasets for robustness evaluation used in our experiments are CIFAR-10 and CIFAR-100 consisting of 50k samples and 10k samples for training and testing, respectively. The image size of both CIFAR datasets are $32\times32$ following the standard protocol in past NAS and noisy label works~\cite{Han2018Co,Liu2018Darts}. We investigate both symmetric and asymmetric label noise. Symmetric noise rates on CIFAR-10 and CIFAR-100 are 20\% and 50\%, and the asymmetric noise rate is 40\%. All results are reported based on the average of the last ten epochs with $\beta=1$. We also run evaluations on a larger  Tiny-ImageNet\footnote{https://tiny-imagenet.herokuapp.com/} with 200 classes and 120k images in total. The image size of Tiny-ImageNet is set to 
$64\times64$ following the setup in~\cite{yu2019does}. Finally, we experiment with the  Clothing1M~\cite{xiao2015learning} which contains real-world noisy labels, 14 categories, and $\sim$1M images ($224\times224$ size). The network is trained for 80 epochs. SGD with momentum is used to optimize the model with the learning rate set to 0.1. Remaining parameters follow the setup for CIFAR. 

\begin{figure}[t]
    \centering
     \includegraphics[width=0.51\textwidth]{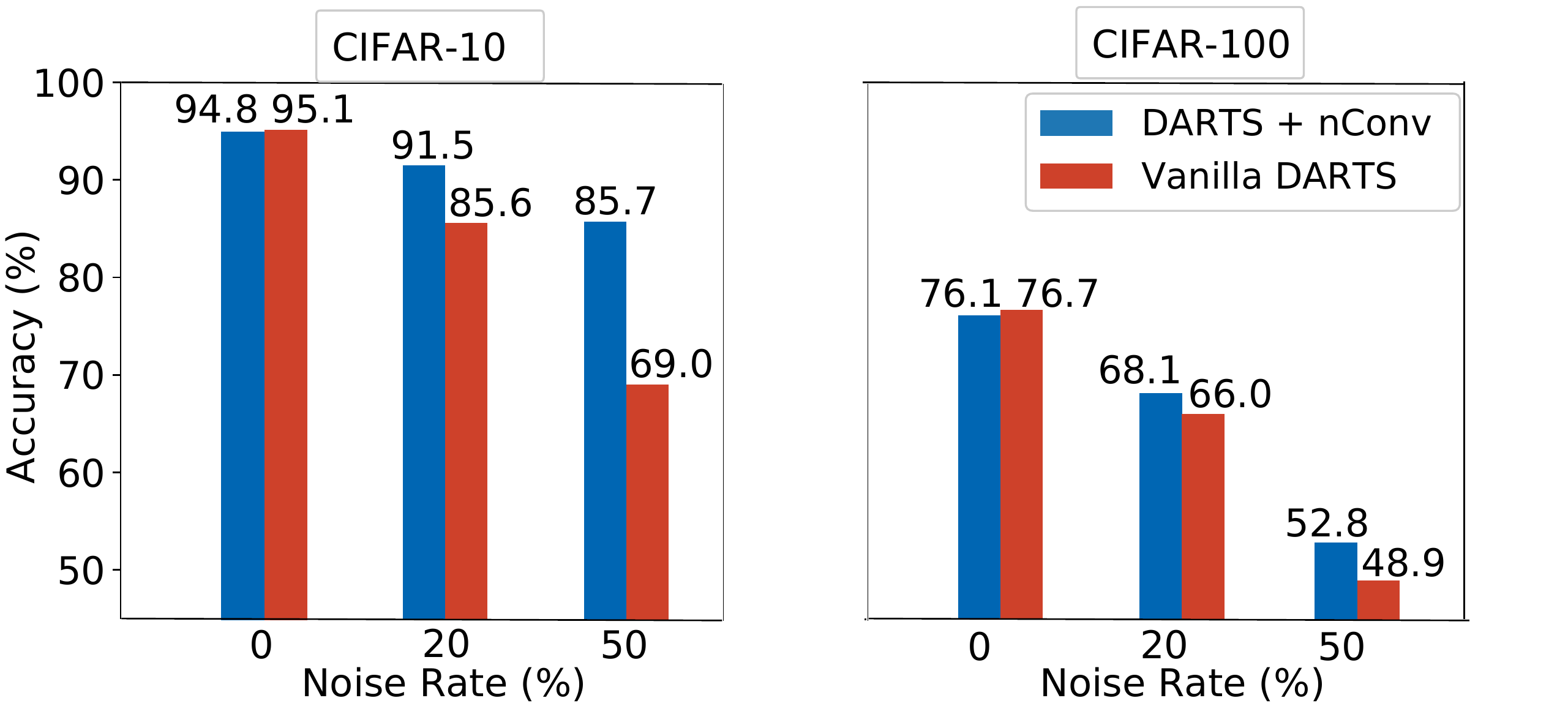}
    \caption{ Neural architecture search under noisy labels with DARTS and DARTS + nConv on CIFAR-10 and CIFAR-100. The comparison uses three protocols: clean, 20\%, and 50\% symmetric label noise.
    }
    \label{fig:compare_darts_performance}
\end{figure}

\subsection{Comparison with Vanilla DARTS}
Below, we compare the performance of Vanilla DARTS and our DARTS with the nConv operator (DARTS + nConv) to study the improvement of the latter when learning in the presence of label noise. We use DARTS with following operators $\mathcal{O}$: $3\times3$ separable convolution, $3\times3$ dilated convolution,  $5\times5$ separable convolution, $5\times5$ dilated convolution, $3 \times 3$ max pooling, $3 \times 3$ average pooling, and the identity. We use the same setup (\eg optimizer parameters/batch size) for both techniques to search and evaluate the found architecture. The search/evaluation phases use the same label noise rates (\eg, searching on CIFAR-10 with 50\% noise and evaluating the found architecture on CIFAR-100 with 50\% noise). 
 Fig.~\ref{fig:compare_darts_performance} shows that DARTS + nConv achieves superior performance compared to Vanilla DARTS.

\begin{table}[h]
    \centering
    \resizebox{.42\textwidth}{!}{
    \Large\addtolength{\tabcolsep}{-.5pt}
    \begin{tabular}{c|c|c|c}
        \hline
          \textbf{Method} &{\textbf{Backbone}}  
           &\textbf{20\%-Sym} &\textbf{50\%-Sym}  \\
         \hline
         F-Correction~\cite{patrini2017making}  &\multirow{4}{2cm}{ \parbox{2cm}{\centering Conv-9 \\(Size: 4.4M)}} & $84.6$ & $59.8$  \\
          Decoupling~\cite{malach2017decoupling}  &  & $80.4$ & $51.5$  \\
         MentorNet~\cite{jiang2018mentornet} &  & $80.8$ & $71.1$  \\
         Co-Teaching~\cite{Han2018Co} & & $82.3$  & $74.0$  \\
         JoCoR~\cite{wei2020combating} &  & $85.7$ & $79.41$ \\
          \hline
         Cross Entropy &\multirow{6}{2.5cm}{\parbox{2.5cm}{\centering ResNet-18 \\(Size: 11.2M)}} &$85.6 $ & $57.8 $   \\
         F-Correction~\cite{patrini2017making}  & & $83.1$ & $59.4$  \\
        Decoupling~\cite{malach2017decoupling}  & & $79.9$ & $52.2$ \\
         MentorNet~\cite{jiang2018mentornet}  & & $80.5$ & $70.7$  \\
         Co-Teaching~\cite{Han2018Co} & & $82.4$  & $72.8$ \\
          $T$ Revision~\cite{xia2019anchor} &  & $89.6$ & $83.4$ \\
          \hline
         \textbf{DARTS + nConv}  &Auto (Size: 1.2M) & $\mathbf{91.4}$ & $\mathbf{85.7}$  \\
         \hline
    \end{tabular}
    }
     \caption{CIFAR-10 test acc. (\%) using ResNet and Conv-9.}
    \label{tab:symmetricnoise_results_cifar10}
\end{table}

\begin{table}[h]
    \centering
    \resizebox{.4\textwidth}{!}{
    \Large\addtolength{\tabcolsep}{-.5pt}
    \begin{tabular}{c|c|c|c}
        \hline
         \textbf{Method} &{\textbf{Backbone}}  
           &\textbf{20\%-Sym} &\textbf{50\%-Sym}  \\
         \hline
         F-Correction~\cite{patrini2017making}  &\multirow{4}{2cm}{ \parbox{2cm}{\centering Conv-9 \\(Size: 4.4M)}} &$61.9$ & $41.0$  \\
          Decoupling~\cite{malach2017decoupling}  &  &$ 44.5$ & $25.8$ \\
         MentorNet~\cite{jiang2018mentornet} &   &$52.1$ & $39.0$  \\
         Co-Teaching~\cite{Han2018Co} & &$54.2$  & $41.4$ \\
         JoCoR~\cite{wei2020combating} &  & $53.0$ & $43.5$ \\
          \hline
         Cross Entropy &\multirow{6}{2.5cm}{\parbox{2.5cm}{\centering ResNet-34 \\(Size: 20.1M)}}  & $61.2$ &$41.2$  \\
         F-Correction~\cite{patrini2017making}  & &$61.4$ & $37.3$ \\
        Decoupling~\cite{malach2017decoupling}  & &$52.1$ & $38.5$ \\
         MentorNet~\cite{jiang2018mentornet}  & & $80.5$ & $70.7$  \\
         Co-Teaching~\cite{Han2018Co} & &$54.2$  & $41.4$  \\
          $T$ Revision~\cite{xia2019anchor} & & $65.4$ & $50.5$ \\
          \hline
         \textbf{DARTS + nConv}  &Auto (Size: 1.2M) & $\mathbf{68.1}$ & $\mathbf{52.8}$  \\
         \hline
    \end{tabular}
    }
     \caption{CIFAR-100 test acc. (\%) using ResNet and Conv-9.}
    \label{tab:symmetricnoise_results}
\end{table}

\begin{table}[h]
    \centering
    \resizebox{.35\textwidth}{!}{
    \small\addtolength{\tabcolsep}{-.1pt}
    \begin{tabular}{c|c|c}
    \hline
         \textbf{Method} &{\textbf{Backbone}} & \textbf{Accuracy (\%)}\\
         \hline
         Cross-entropy &\multirow{2}{2cm}{ \centering ResNet-18 \\(Size: 11.2M)} &72.3\\
         F-Correction~\cite{patrini2017making} & & 83.1\\
         \hline
         \textbf{DARTS + nConv} &Auto (Size: 1.2M) & $\mathbf{89.5}$\\
    \hline     
    \end{tabular}}
      \caption{Testing acc. (\%) with 40\% asymmetric noise on CIFAR-10.}
    \label{tab:assymetricnoise_results}
\end{table}

\begin{table}[h]
    \centering
    \resizebox{.4\textwidth}{!}{
    \small\addtolength{\tabcolsep}{-.2pt}
    \begin{tabular}{c|c|c|c}
    \hline
         \textbf{Method} &{\textbf{Backbone}} &\textbf{ 20\%-Sym} & \textbf{50\%-Sym}\\
         \hline
         Decoupling~\cite{malach2017decoupling} &\multirow{3}{2cm}{ \parbox{2cm}{\centering ResNet-18 \\(Size: 11.2M)}} &36.3 &22.6\\
         F-Correction~\cite{patrini2017making} & &44.4 &32.8\\
         MentorNet~\cite{jiang2018mentornet}& &45.5 &35.5\\
         Co-teaching~\cite{Han2018Co}& &45.6 &37.1\\
         Co-teaching+~\cite{yu2019does}& &47.7 &41.2\\
         \hline
         \textbf{DARTS + nConv} &Auto (Size: 1.2M) &$\mathbf{50.6}$ & $\mathbf{45.7}$\\
    \hline     
    \end{tabular}}
  \caption{Testing acc. (\%) with 20\% and 50\% symmetric noise on Tiny-ImageNet.}
    \label{tab:tinyiimagenet_results}
    \vspace{-0.4cm}
\end{table}

\begin{figure*}[t]
    \centering
 \subfloat[]{
 \includegraphics[width=0.22\textwidth]{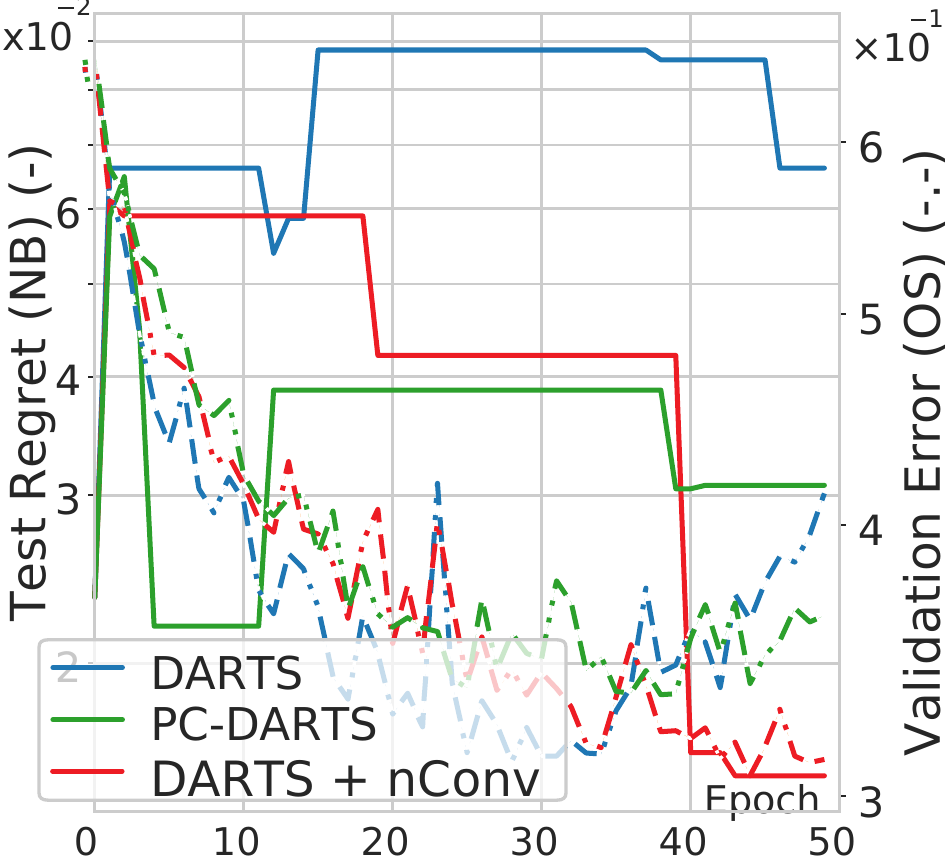}}\quad
  \subfloat[]{
        \includegraphics[width=0.22\textwidth]{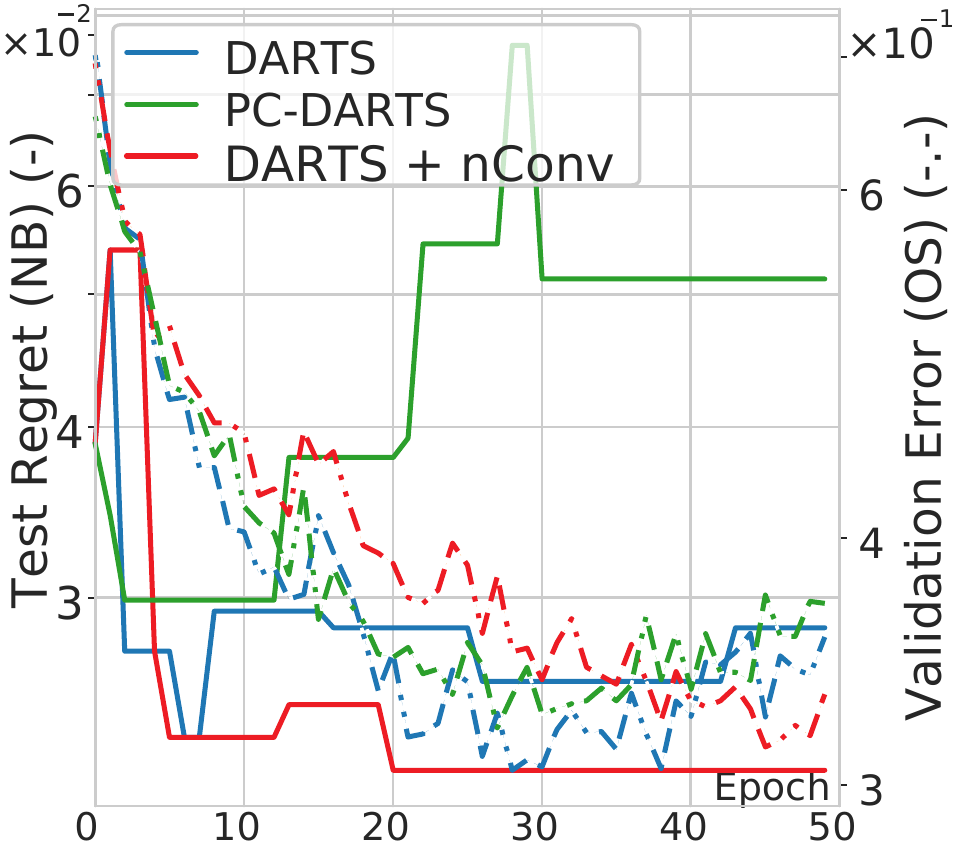}}\quad
  \subfloat[]{
        \includegraphics[width=0.22\textwidth]{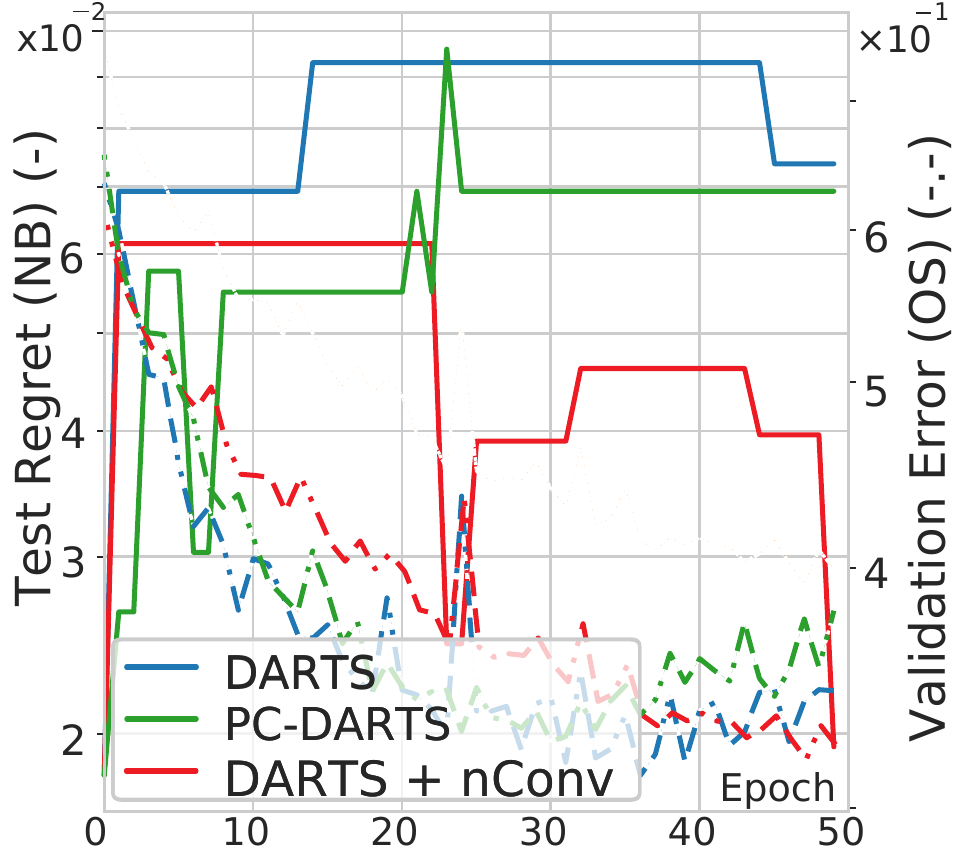}}\quad
    \subfloat[]{\includegraphics[width=0.23\textwidth]{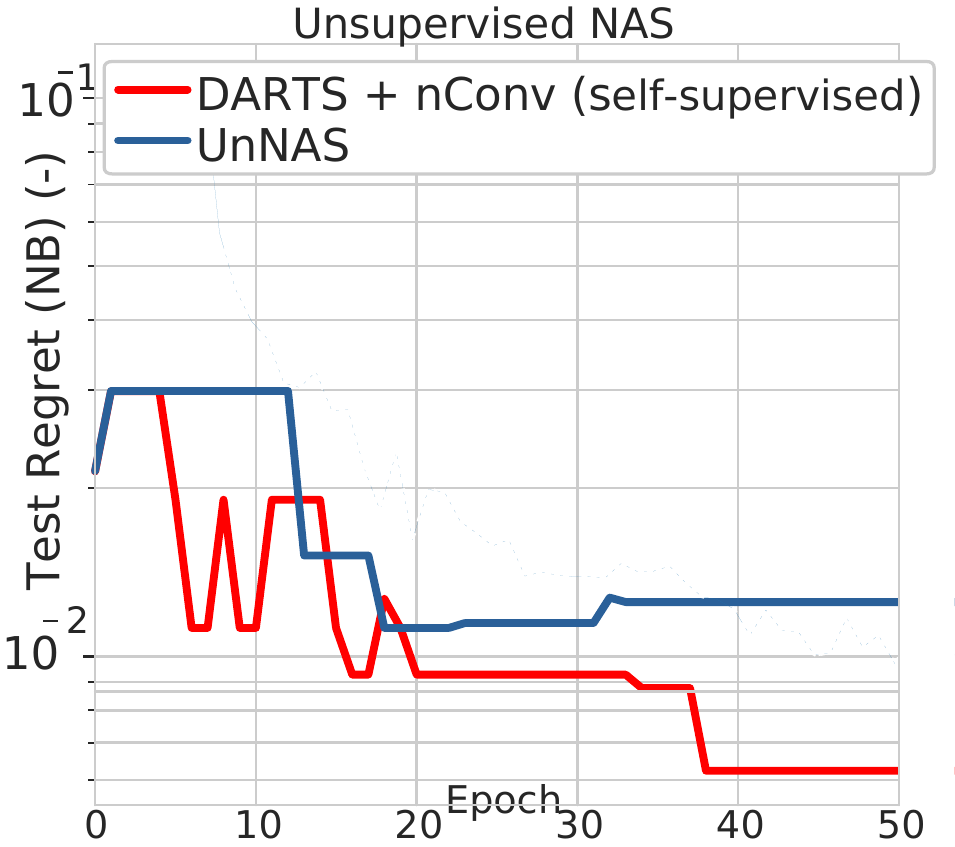}}
    \caption{Test regret of NAS benchmark 1shot1 on CIFAR-10 (searching with 50\% symmetric noise) using three search spaces: (a) search space 1, (b) search space 2, and (c) search space 3. (d) Test regret of the self-supervised setting using search space 2.
    }
    \label{fig:benchmark}
\end{figure*}

\subsection{Comparison with the state of the art for training under label noise}
We compare our method to multiple baselines that are designed to tackle problems resulting from label noise. The baselines we compare to, use three different backbones: Conv-9~\cite{Han2018Co}, ResNet-18, and ResNet-34~\cite{He2016ResNet}.  Table~\ref{tab:symmetricnoise_results} shows the results on CIFAR-10 and CIFAR-100 in the presence of 20\% as well as 50\% symmetric noise. Our proposed method with nConv outperforms T-Revision~\cite{xia2019anchor} on all protocols by a substantial margin of 2\%.  In addition, Table~\ref{tab:assymetricnoise_results} shows that our proposed method consistently performs better than the state of the art \ie, joint-optimization~\cite{tanaka2018joint}, PENCIL~\cite{yi2019probabilistic}, and F-Correction~\cite{patrini2017making}. Note that our approach does not need any modifications of the loss function to train the model under symmetric or asymmetric noise.

We further evaluate the performance on challenging datasets, namely Tiny-ImageNet and Clothing1M~\cite{xiao2015learning}. Note that our evaluation does not use pre-trained neural networks for both datasets. As shown in Table~\ref{tab:tinyiimagenet_results}, DARTS + nConv also outperforms Co-teaching+~\cite{yu2019does}, Co-teaching~\cite{Han2018Co}, MentorNet~\cite{jiang2018mentornet}, and F-Correction~\cite{patrini2017making}. When the noise rate increases from 20\% to 50\%, the performance gap is widened. As an example, the performance gap increases by 3\% when comparing our method with Co-teaching~\cite{Han2018Co}. Moreover, Table~\ref{tab:clothing1m} shows that our method  outperforms state-of-the-art methods on the Clothing1M dataset by 1.3\%. 

Our proposed method outperforms current state of the art  on all datasets irrespective of noise rates  while substantially reducing the number of parameters compared to Conv-9, ResNet-18, and ResNet34 backbones. The searched architecture has fewer parameters (size reduced by 10-20$\times$) compared to ResNet backbones. Finally, our approach maintains only  one neural network while previous methods (\eg, Decoupling~\cite{malach2017decoupling} and Co-teaching~\cite{Han2018Co}) train two networks simultaneously.  

\begin{table}[h]
    \centering
    \resizebox{.49\textwidth}{!}{
    \Large\addtolength{\tabcolsep}{-.5pt}
    \begin{tabular}{c|c|c|c|c}
        \hline
         \textbf{Dataset} & \textbf{DARTS + nConv} & \textbf{Co-Teaching}~\cite{Han2018Co} & \textbf{Decoupling}~\cite{malach2017decoupling}  & \textbf{F-Correction}~\cite{patrini2017making} \\
         \hline
         Clothing1M &\textbf{69.8} &68.5 &67.3 &65.4 \\
         \hline
    \end{tabular}
    }
        \caption{Test acc. on the Clothing1M dataset. The prior methods use ResNet-18.}
        \vspace{-0.2cm}
    \label{tab:clothing1m}
\end{table}

\subsection{NAS benchmark evaluation}
Below, we examine the capability of DARTS to obtain high performance architectures in the presence of label noise.  We compare our method with Vanilla DARTS and PC-DARTS~\cite{Xu2020PC-DARTS} using NAS-benchmark 1Shot1~\cite{Zela2020NASBench1Shot1}. There are three different search spaces (1,2, and 3) used in our evaluation which differ by the number of nodes and node parents. The search space 3 in this benchmark is the largest search space ($\sim$360,000 architectures) among  NAS benchmarks. CIFAR-10 is used to search the architecture with 50\% symmetric noise.  Fig.~\ref{fig:benchmark} shows that our method does not overfit to the noisy labels and performs better (lower test regret) compared to Vanilla DARTS and PC-DARTS. Fig

\section{Discussion}
\label{sec:discussion}

In our experiments, we have observed that reducing the mutual information by adjusting the noise variance (following the ELBO derivations) according to data helps us overcome overfitting to the label noise. 
The benefit of our structural approach is that we train a neural network in the presence of label noise with fewer hyper-parameters \eg, no noise rate~\cite{Han2018Co} or a bootstrapping parameter~\cite{reed2015training}. Moreover, a standard training loss (\eg, cross-entropy) can be used directly without any modifications (\eg, multiplying with a certain weight or hinge loss) for both label noise cases.

In the recent NAS work, Liu \etal~\cite{liu2020labels} proposes a self-supervised method so-called UnNAS to search an optimal  architecture. However, the labels of pretext tasks (\eg, rotation, scaling, permutations \cite{zhang2020few,arl}) suffer from the noise depending on the initial state of the input. Even unsupervised, contrastive and hallucinating methods \cite{ssgc,refine,coles,i3d_halluc,i3d_halluc2} suffer from the diffusion, negative sampling noise or quantization noise. Yet, NAS using our approach  is able to find a better architecture compared to UnNAS~\cite{liu2020labels}. Fig.~\ref{fig:benchmark} (d) shows that our method outperforms the result of UnNAS by achieving lower test regret of NAS benchmark 1shot1.

\section{Conclusions}
\label{sec:conclusions}
We have presented a new operator coined \textbf{nConv} to search and train a robust neural network in the presence of label noise. 
We have shown that label noise harms the performance of Vanilla DARTS while DARTS + nConv performs robustly on problems with noisy labels.   
The proposed operator has a minimum amount of overhead compared to the existing operators in the search space (\eg, convolution operations in Vanilla DARTS). 
As changes are applied to the neural network structure, mechanisms such as sample selection, modified loss functions or transition matrices are not needed by our approach. 
Our empirical results also show that we can design a  neural network with fewer parameters compared to ResNet backbones, without the loss of accuracy.

{\small
\bibliographystyle{ieee_fullname}
\bibliography{egbib}
}






In this supplementary material, we provide the implementation details of our method, additional results, and our training strategy.

\section{Derivation}
Recall in the main paper we have $I(Z; \BLUE{\hat{Y}}, \Vec{\phi})$ and $I(Z;\RED{\widetilde{Y}},\Vec{\phi})$ with the lower bound defined as follows:

\begin{align}
\begin{split}
I(Z;\BLUE{\hat{Y}}, \Vec{\phi}) +  I(Z;{\RED{\widetilde{Y}}}, \Vec{\phi}) - \beta I(X;Z, \Vec{\phi}) \geq \\
\int p(\Vec{x})p(\BLUE{\Vec{\hat{y}}} \vert \Vec{x}; \Vec{\phi})p(\Vec{z} \vert \Vec{x}; \Vec{\phi}) \log \bar{p}(\BLUE{\Vec{\hat{y}}} \vert \Vec{z}; \Vec{\phi}) \;d\Vec{x} \;d\Vec{\hat{y}}\; d\Vec{z} \\
+ \int p(\Vec{x})p(\RED{\Vec{\widetilde{y}}} \vert \Vec{x}; \Vec{\phi})p(\Vec{z} \vert \Vec{x}; \Vec{\phi})
\log \bar{p}(\RED{\Vec{\widetilde{y}}} \vert \Vec{z}; \Vec{\phi}) \;d\Vec{x} \;d\Vec{\widetilde{y}}\; d\Vec{z} \\
- \beta\int p(\Vec{x})p(\Vec{z}|\Vec{x}; \Vec{\phi})\mathrm{log}\frac{p(\Vec{z} \vert \Vec{x}; \Vec{\phi})}{p(\Vec{z}\vert \Vec{\phi})} \;d\Vec{x}\; d\Vec{z}
= \mathcal{L}. 
\end{split}
\label{eq:lowerbound_suppmat}
\end{align}
The lower bound $\mathcal{L}$ in Eq.~\ref{eq:lowerbound_suppmat} can be approximated using the formulation for variational inference $q(\Vec{z}|\Vec{x};\Vec{\phi})$: 

\begin{align}
\begin{split}
\mathcal{L} = 
   \underbrace{\int p(\Vec{x})p(\BLUE{\Vec{\hat{y}}} \vert \Vec{x}; \Vec{\phi})p(\Vec{z} \vert \Vec{x}; \Vec{\phi}) \log \bar{p}(\BLUE{\Vec{\hat{y}}} \vert \Vec{z}; \Vec{\phi}) \;d\Vec{x} \;d\Vec{\hat{y}}\; d\Vec{z}}_{ \approx \mathbb{E}_{\Vec{z} \sim q(\Vec{z} \vert \Vec{x}; \Vec{\phi})}\big{[} \log p(\BLUE{\hat{\Vec{y}}_n} | \Vec{z}) \big{]}}\\
   + \underbrace{\int p(\Vec{x})p(\RED{\Vec{\widetilde{y}}} \vert \Vec{x}; \Vec{\phi})p(\Vec{z} \vert \Vec{x}; \Vec{\phi})
\log \bar{p}(\RED{\Vec{\widetilde{y}}} \vert \Vec{z}; \Vec{\phi}) \;d\Vec{x} \;d\Vec{\widetilde{y}}\; d\Vec{z}}_{\approx \mathbb{E}_{\Vec{z} \sim q(\Vec{z} \vert \Vec{x}; \Vec{\phi})}\big{[} \log p(\RED{\widetilde{\Vec{y}}_n} | \Vec{z}) \big{]}} \\
- \underbrace{\beta\int p(\Vec{x})p(\Vec{z}|\Vec{x}; \Vec{\phi})\mathrm{log}\frac{p(\Vec{z} \vert \Vec{x}; \Vec{\phi})}{p(\Vec{z}\vert \Vec{\phi})} \;d\Vec{x}\; d\Vec{z}}_{\approx \beta \text{KL}\big{[} q(\Vec{z}|\Vec{x}_n; \Vec{\phi}) \vert \vert r(\Vec{z}) \big{]}}.
\end{split}
\label{eq:lowerbound_minimize}
\end{align}
The first and second terms in Eq.~\ref{eq:lowerbound_minimize} are the average cross-entropy and the third term is a regularization term. Our objective is to attain the low value of $\mathcal{L}$ but we should avoid minimizing the second term that operates on noisy labels to avoid overfitting to these erroneously labeled datapoints.

To this end, the noise injection by the noise injecting operator (which actually approximates the information bottleneck) acts as a regularization to the loss function.


\begin{figure*}[t]
    \centering
     \subfloat[Train ($\Vec{\mu}=0.0$)]{
    \includegraphics[width=0.33\textwidth]{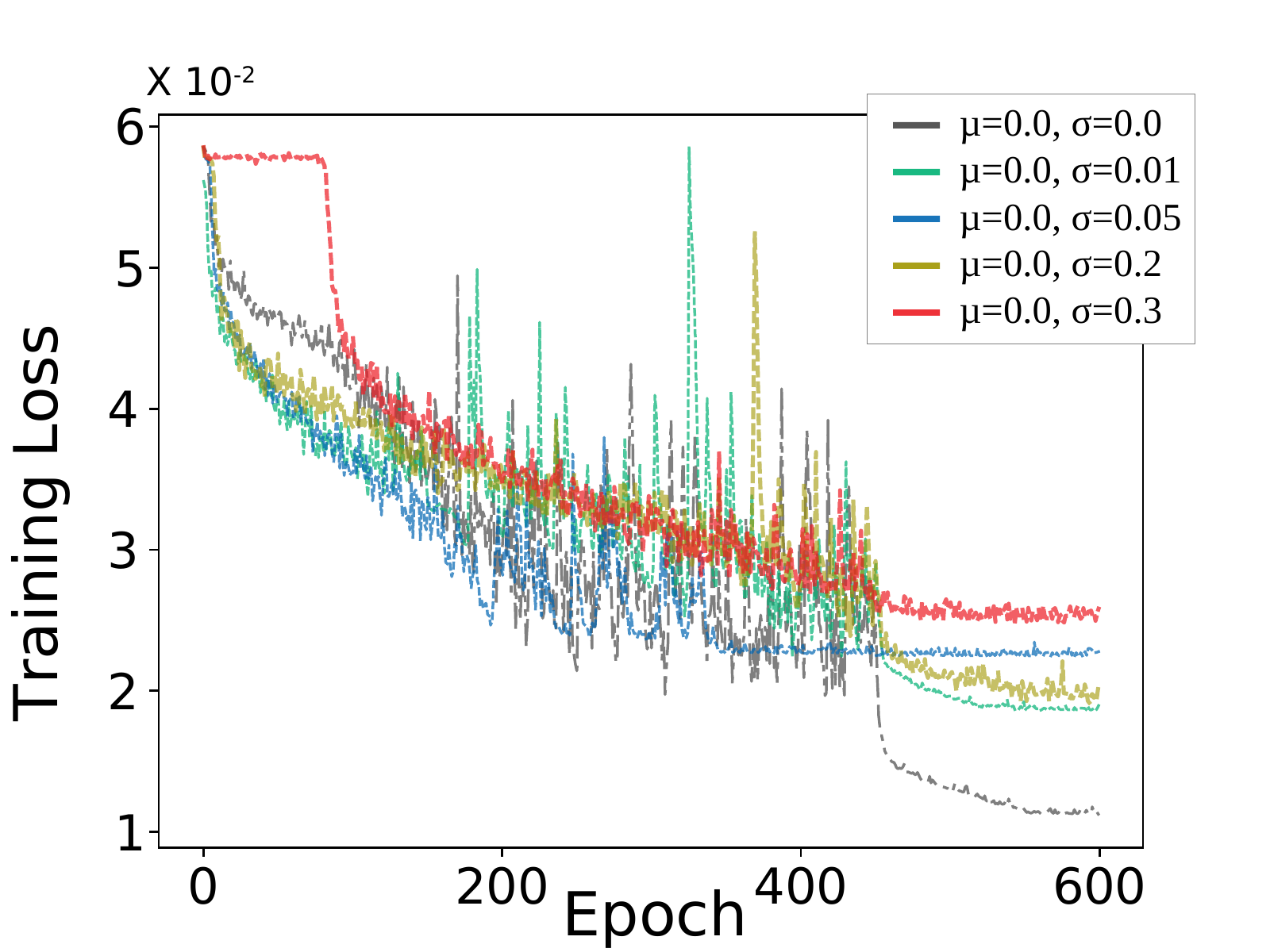}}  \quad
    \subfloat[Validation ($\Vec{\mu}=0.0$)]{
        \includegraphics[width=0.33\textwidth]{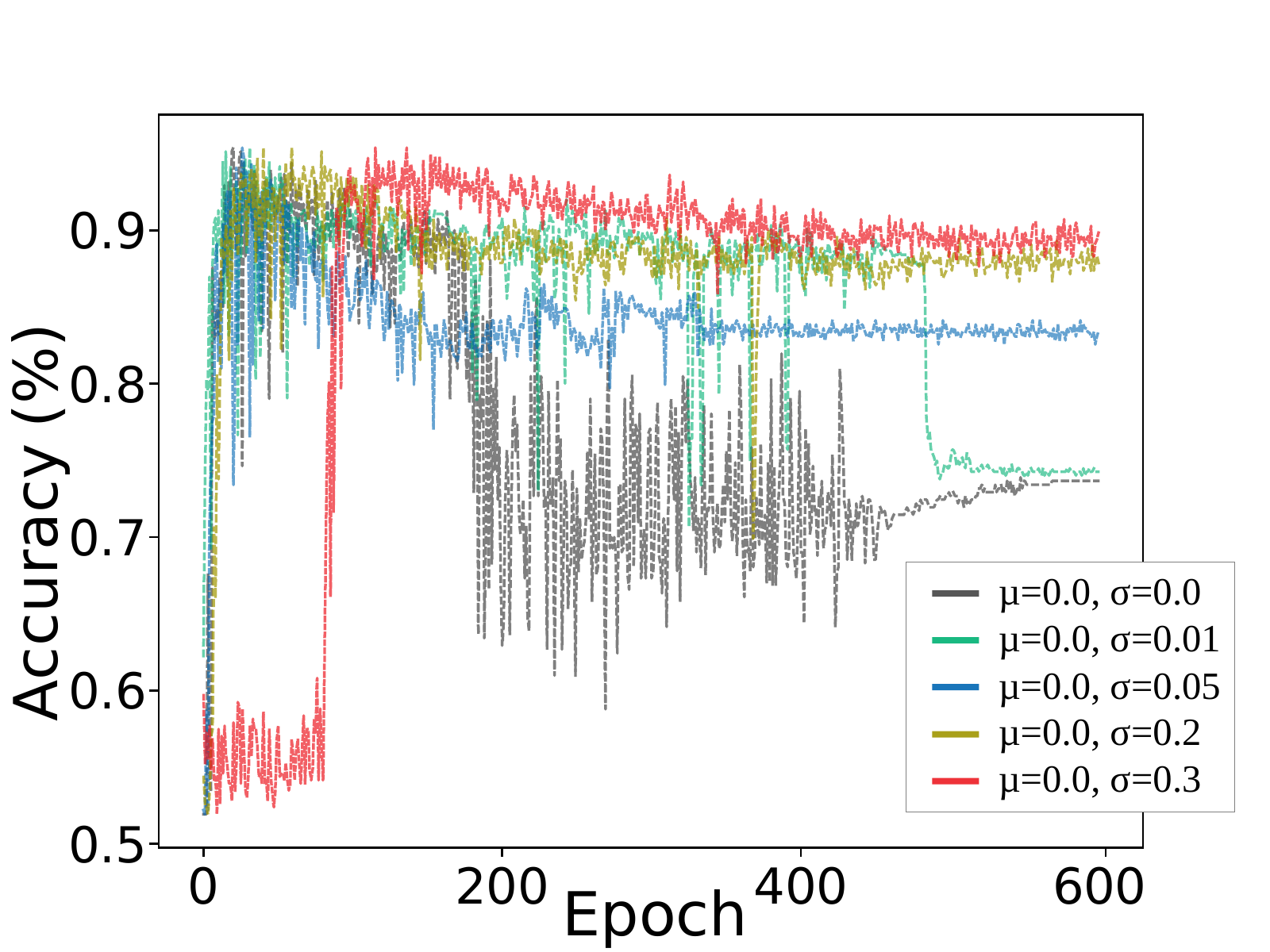}}\\
    \subfloat[Train ($\Vec{\mu}=0.1$)]{
    \includegraphics[width=0.33\textwidth]{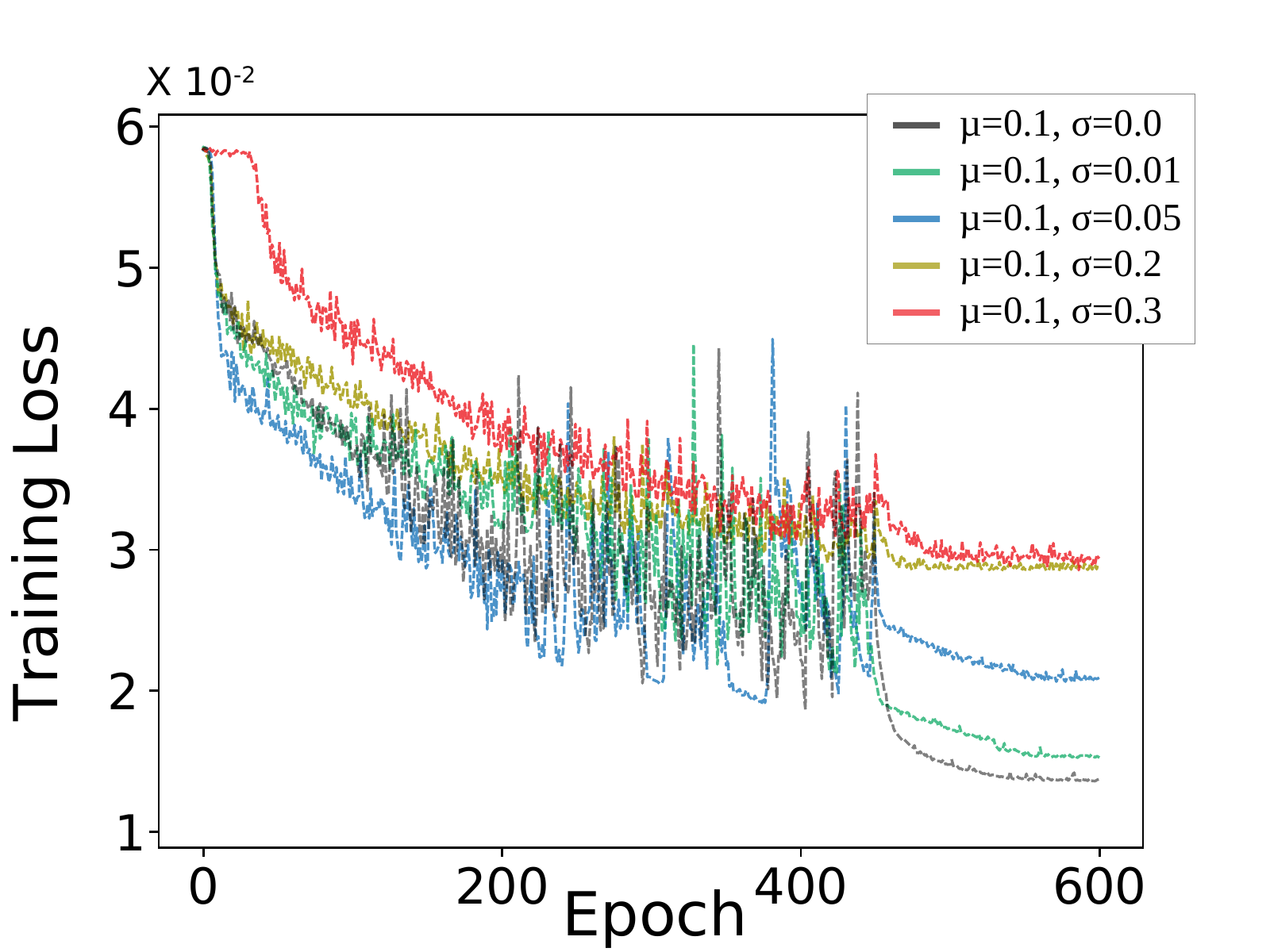}}  \quad
    \subfloat[Validation ($\Vec{\mu}=0.1$)]{
        \includegraphics[width=0.33\textwidth]{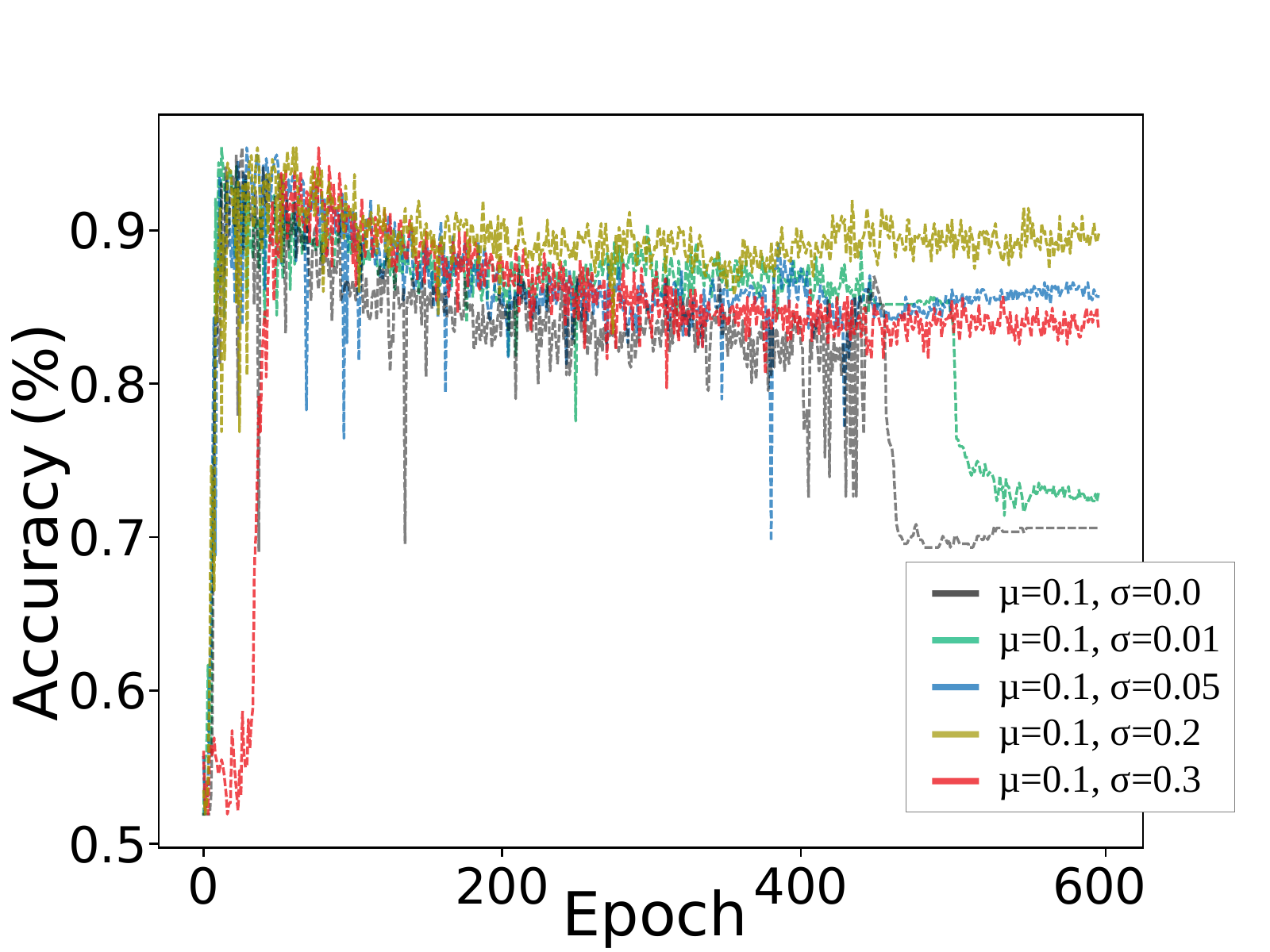}}\\
    \subfloat[Train ($\Vec{\mu}=0.2$)]{
    \includegraphics[width=0.33\textwidth]{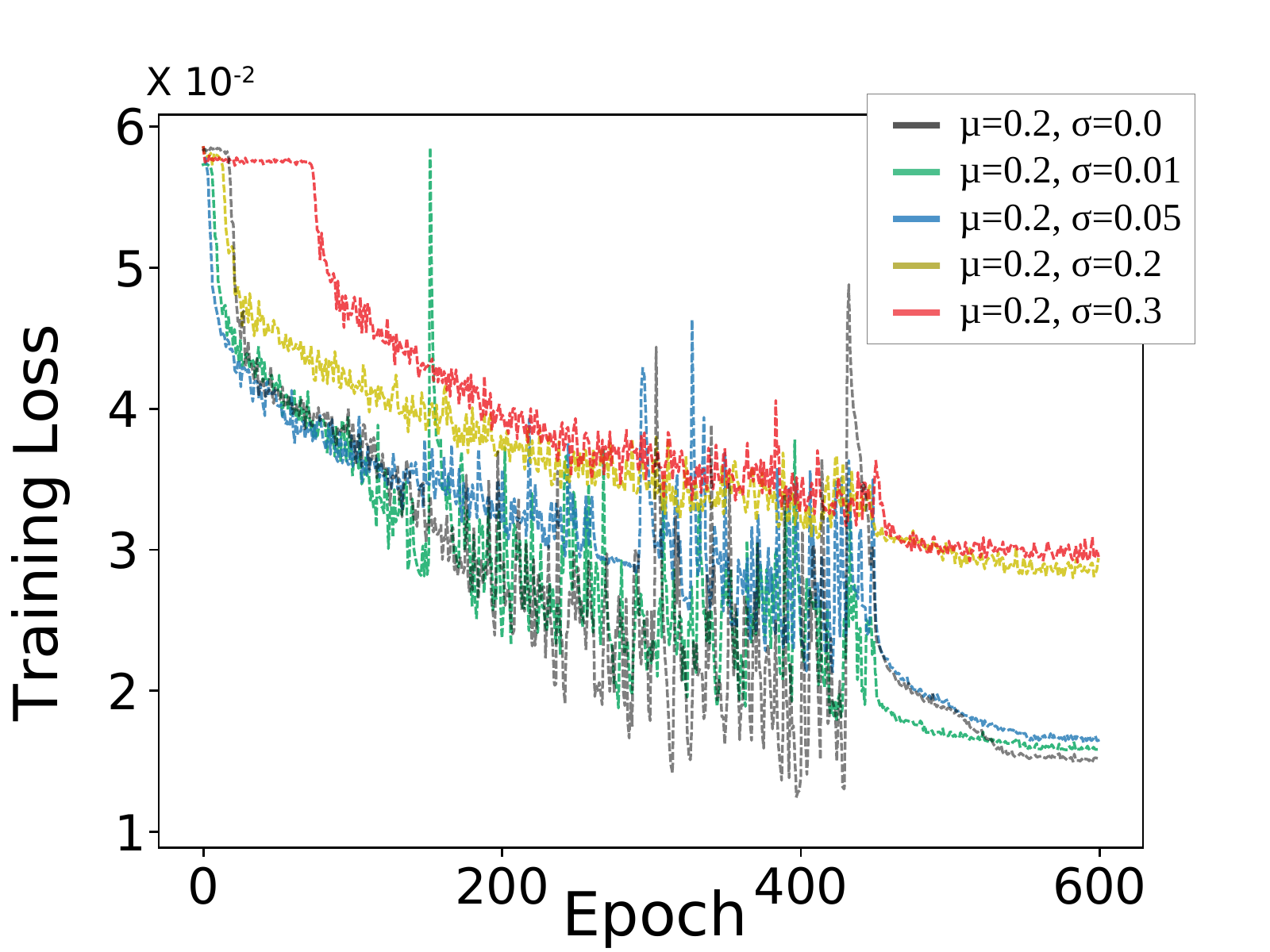}}  \quad
    \subfloat[Validation ($\Vec{\mu}=0.2$)]{
        \includegraphics[width=0.33\textwidth]{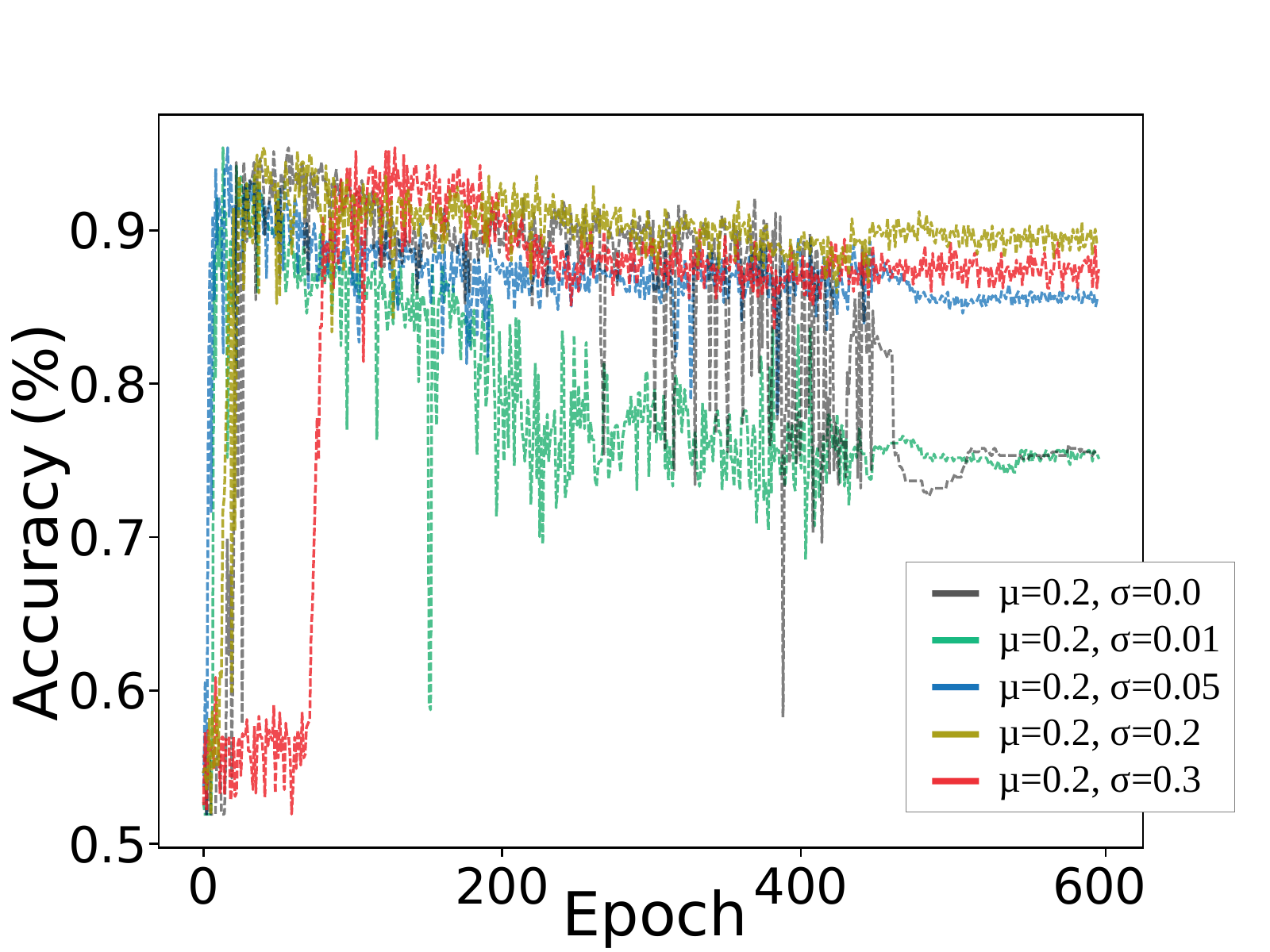}} \\
    \subfloat[Train ($\Vec{\mu}=0.3$)]{
    \includegraphics[width=0.33\textwidth]{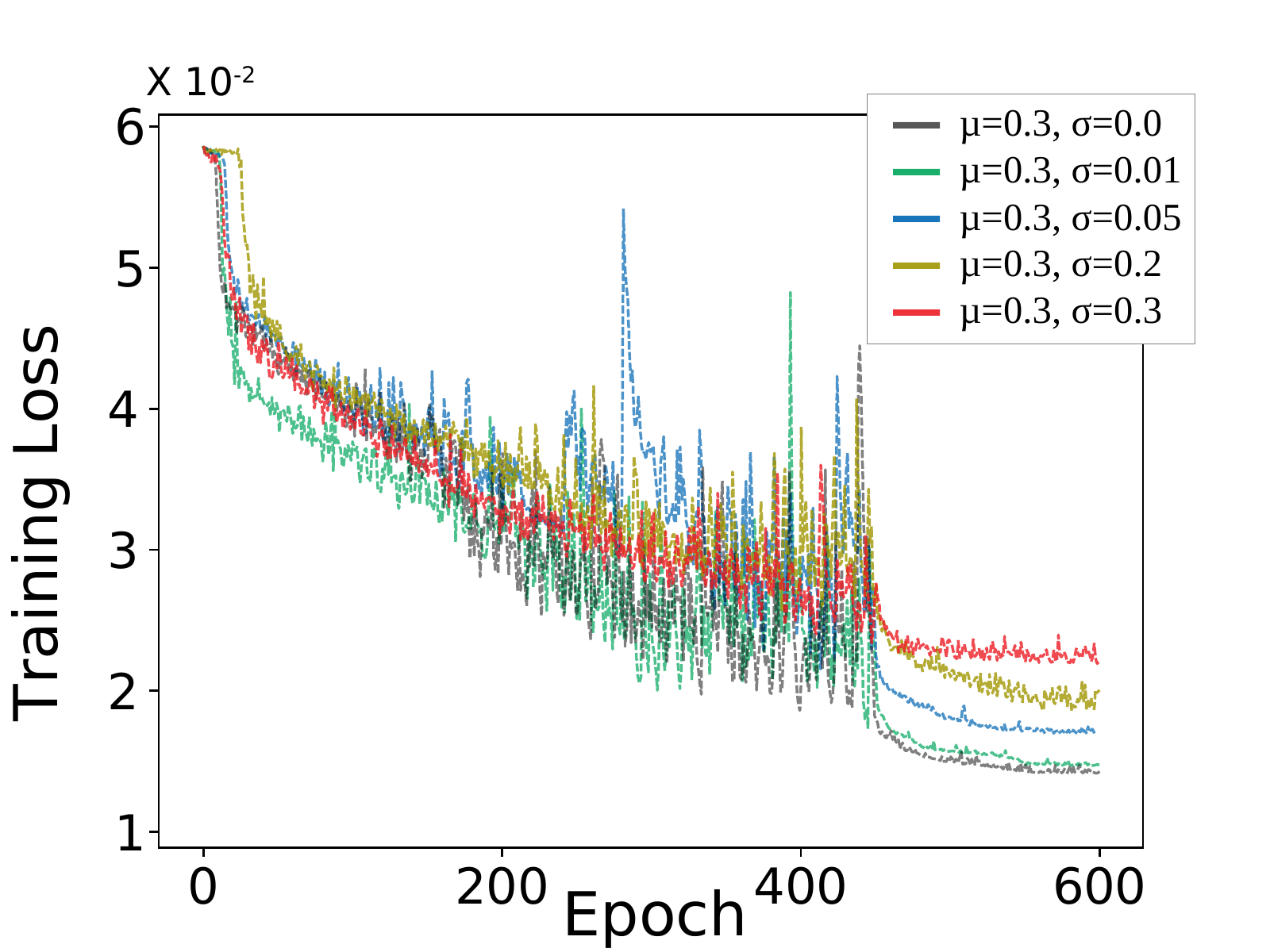}}  \quad
    \subfloat[Validation ($\Vec{\mu}=0.3$)]{
        \includegraphics[width=0.33\textwidth]{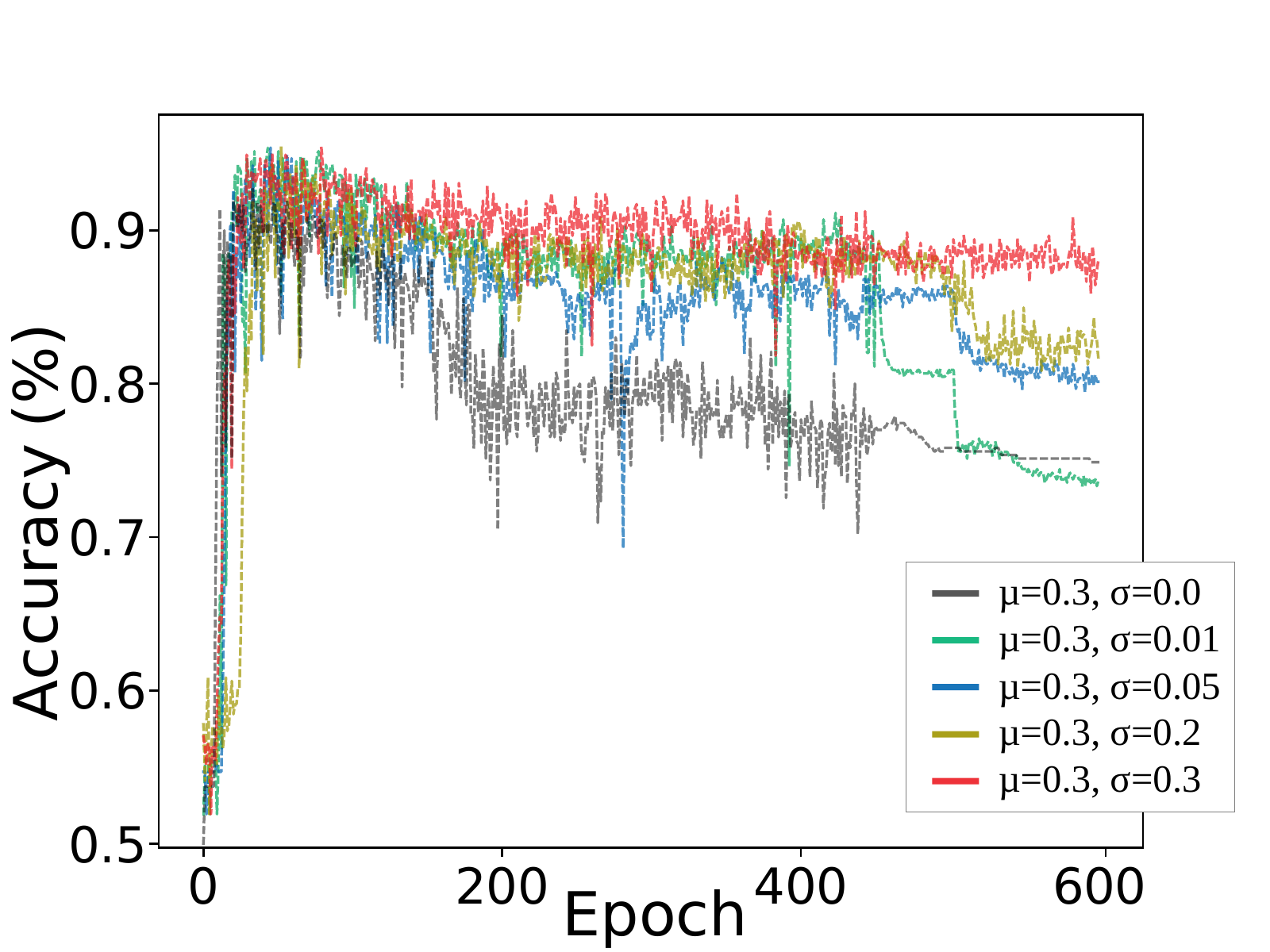}}     
    \caption{Training under 20\% symmetric noise on the toy data by varying the standard deviation according to $\{0.0, 0.01, 0.05, 0.2, 0.3\}$ and the mean according to $\{0.0, 0.1, 0.2, 0.3\}$. }
     \label{fig:train_vary_sigma}
\end{figure*}

\begin{figure*}[t]
    \centering
\subfloat[]{
        \includegraphics[width=0.31\textwidth]{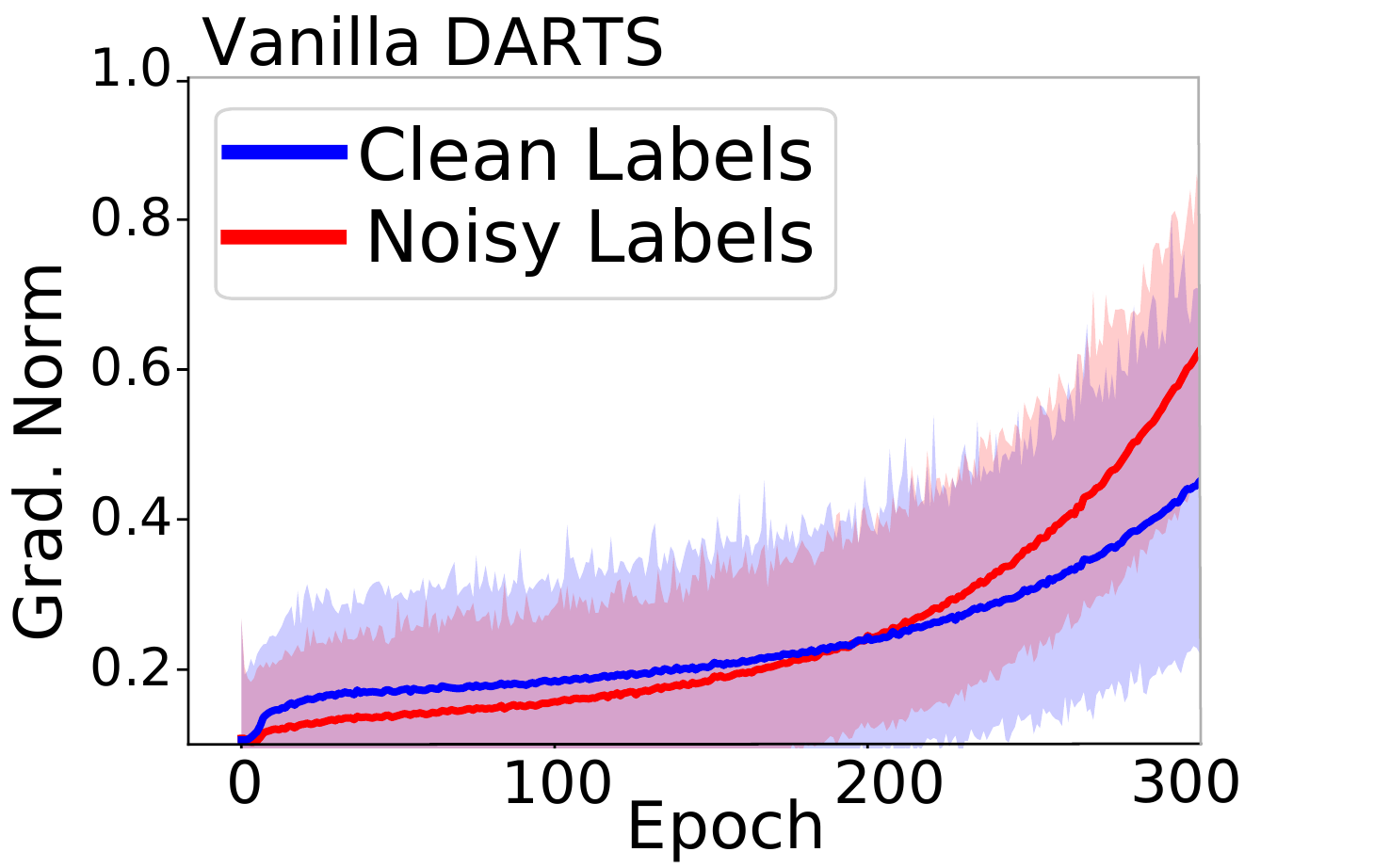}}
\quad \subfloat[]{
    \includegraphics[width=0.31\textwidth]{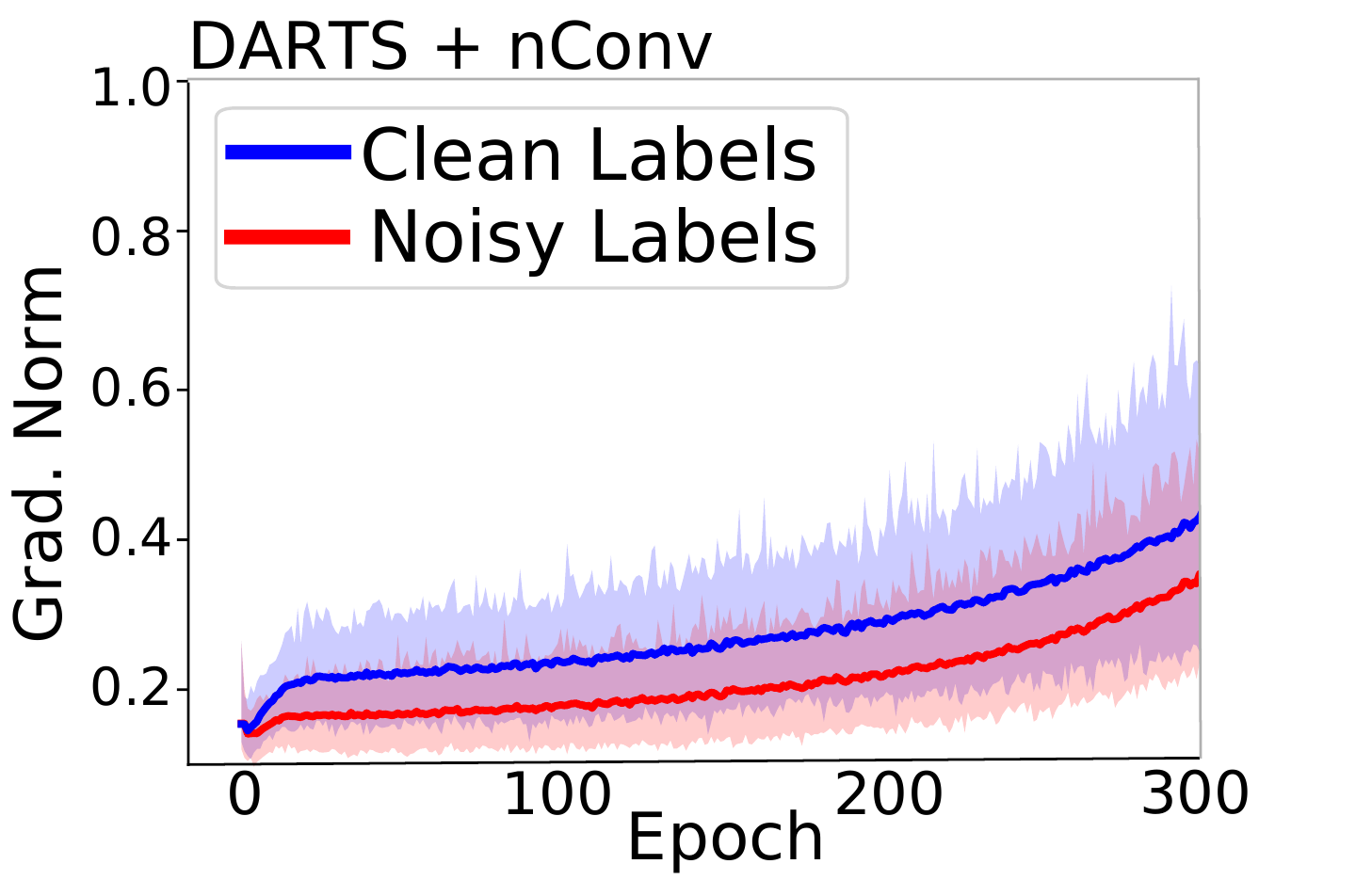}} 
    \quad
    \subfloat[]{
    \includegraphics[width=0.285\textwidth]{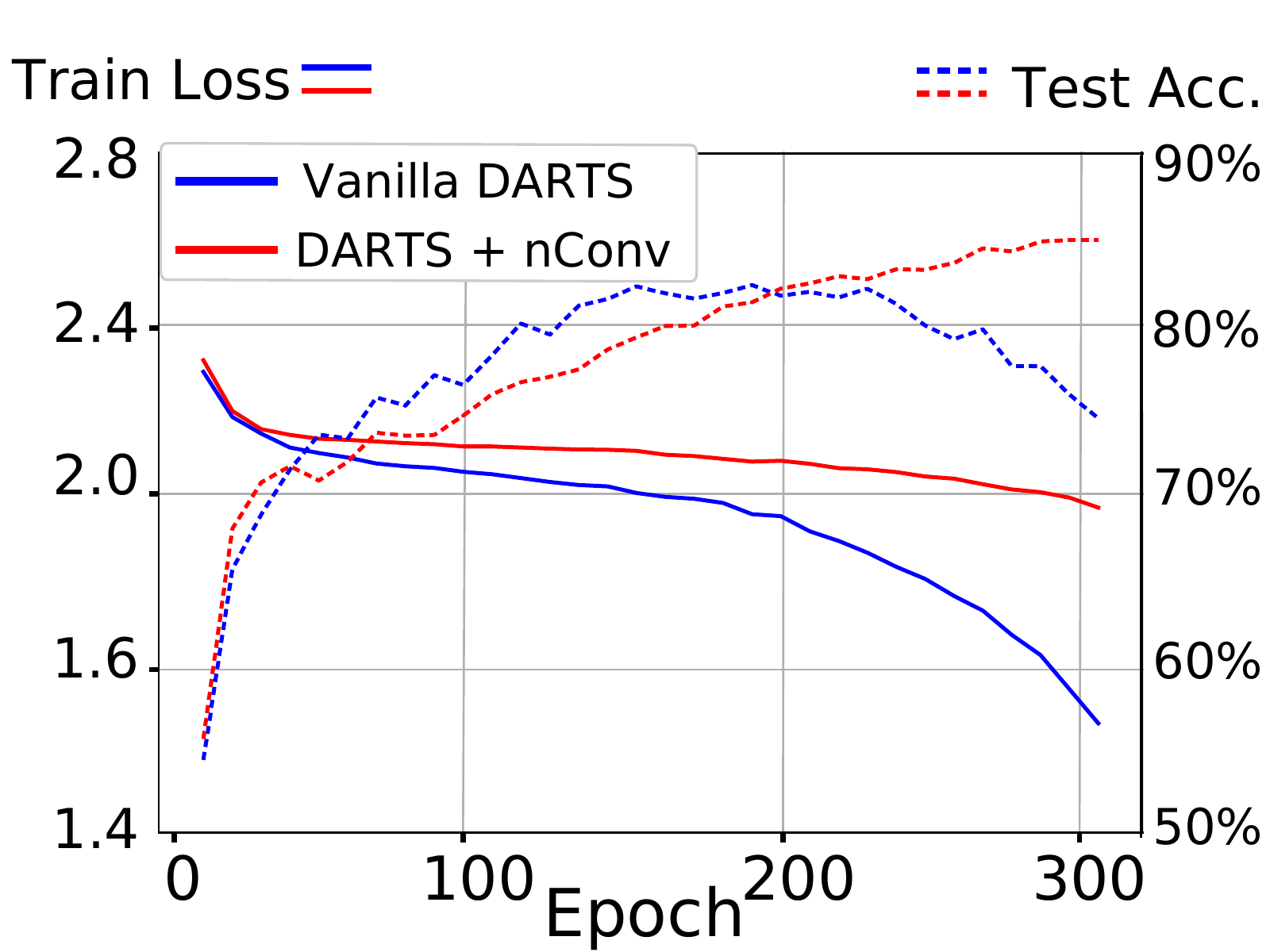}}   
    \caption{ The norm of gradients (the mean with standard deviation) while training the found architectures (evaluation phase) on CIFAR-10 with 50\%-symmetric label noise. (a) Vanilla DARTS is highly impacted by the gradients of samples with noisy labels after 200 epochs. (b) DARTS + nConv maintains the norm of gradients from both (noisy and clean) samples. (c) Train loss and validation accuracy in the evaluation phase of NAS as a function of epoch number.  
    }
    \label{fig:noisegrad}
\end{figure*}

\section{Implementation details}
Our noise injection operator consists of adaptive average pooling, two fully connected layers with a ReLU function baetween them, and a Sigmoid function. The average pooling operation shrinks the height and width $\mathbb{R}^{C\times H \times W}$ of the hidden unit to  $\mathbb{R}^{C \times 1 \times 1}$ on which $\Vec{\sigma} \in \mathbb{R}^{C\times 1 \times 1}$ is learnt with a simple Multi-layer Perceptron (MLP).  Thus, a noise injection module which generates $\Vec{\sigma} \in \mathbb{R}^{C\times 1 \times 1}$, performs a channel-wise variance scaling step to the noise (tensor of size $\mathbb{R}^{C\times H \times W}$) sampled  from a Normal distribution. Subsequently, the scaled noise is directed to the output of the operator. Finally, the output is added with the residual of the input.


There are two types of convolutions in the search space: the dilated convolution and the separable convolution. 
For instance, the separable convolution with 3x3 kernel (\textit{sep\_conv\_3x3}) and the dilated convolution with 3x3 kernel (\textit{dil\_conv\_3x3}) are presented in Tables~\ref{tab:sepconv}  and~\ref{tab:dilconv}, respectively. We also provide examples of kernels given the injected noise in the separable convolution with 3x3 kernel (\textit{sepconv3x3\_noise}) and the dilation convolution with 3x3 kernel (\textit{dilconv3x3\_noise}).

\section{Searching architectures}
\begin{figure*}[t]
    \centering
     \subfloat[Normal cell (20\%-sym)]{
    \includegraphics[width=0.52\textwidth]{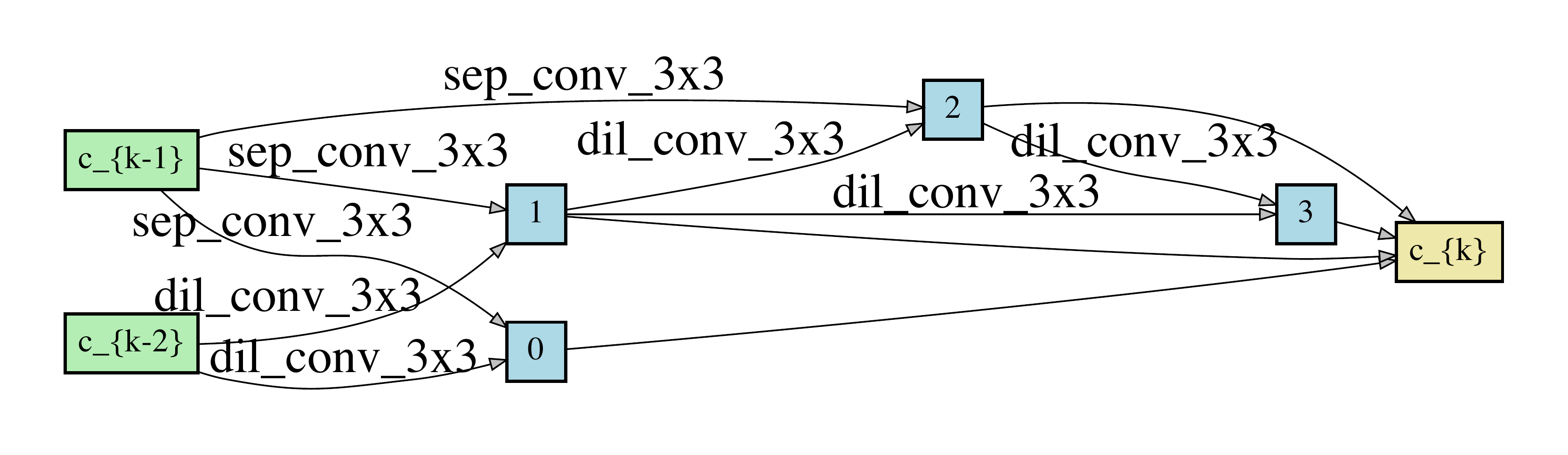}}  
    \subfloat[Reduction cell (20\%-sym)]{
        \includegraphics[width=0.5\textwidth]{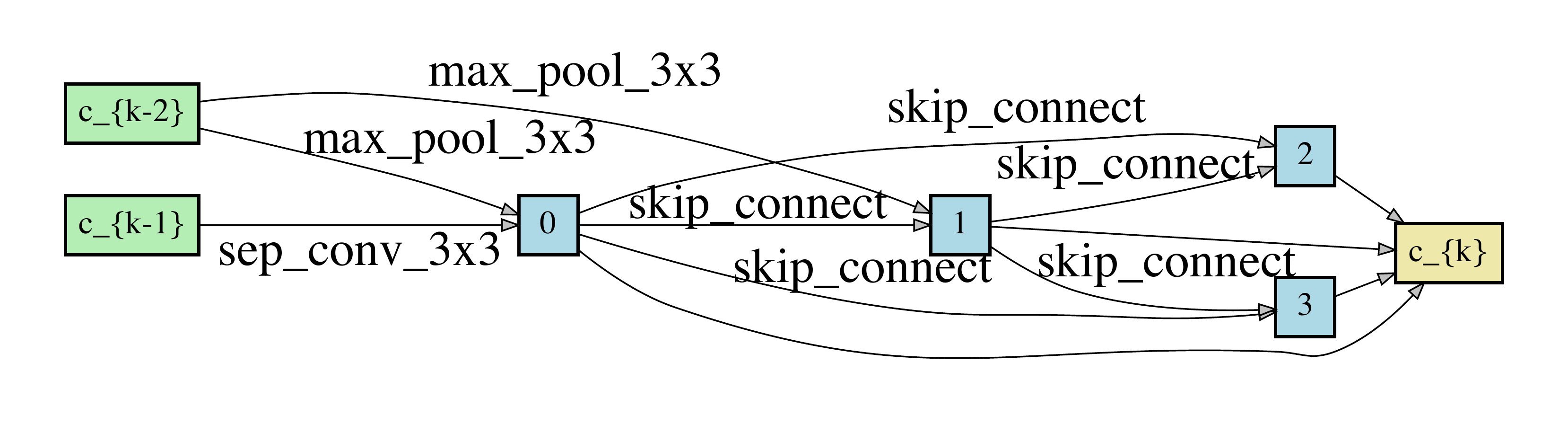}}   
    
     \subfloat[Normal cell  (50\%-sym)]{\includegraphics[width=0.5\textwidth]{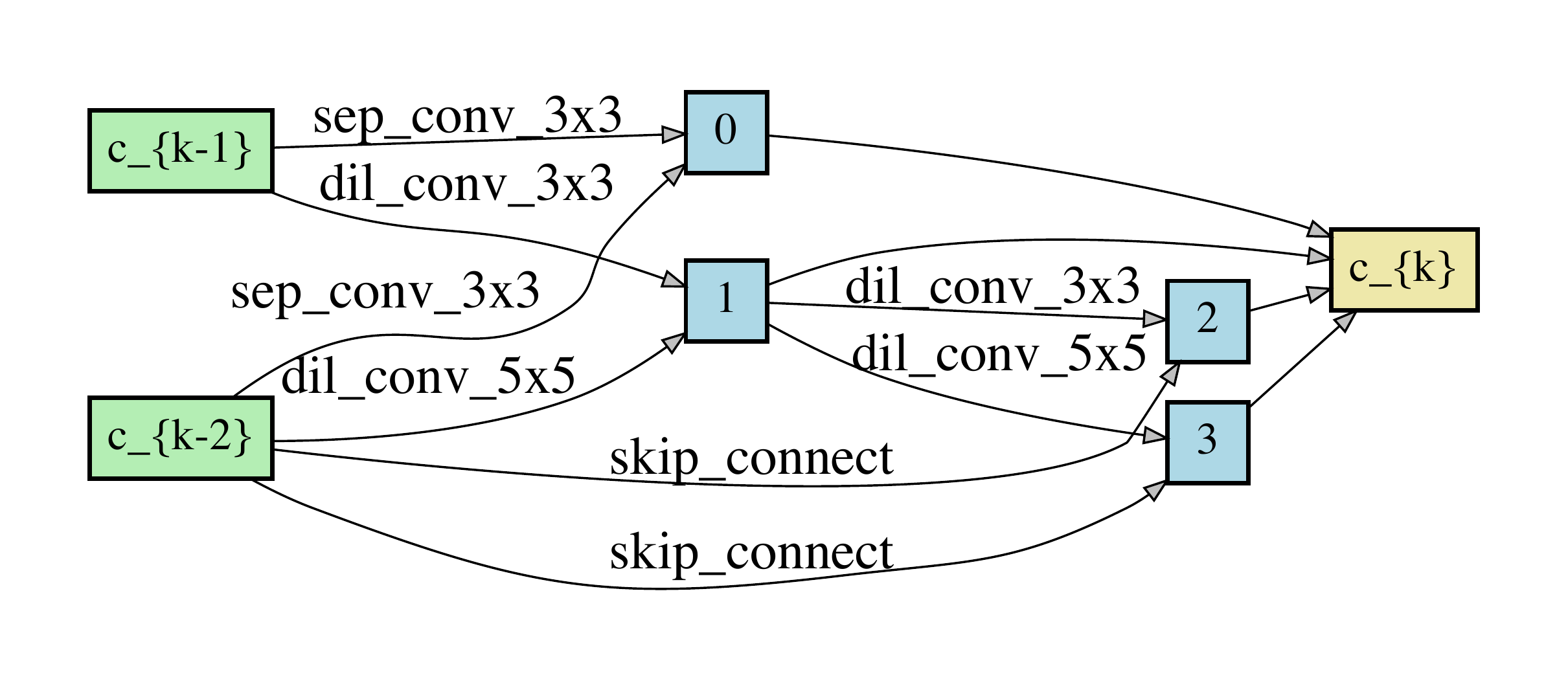}}  
    \subfloat[Reduction cell (50\%-sym)]{
        \includegraphics[width=0.5\textwidth]{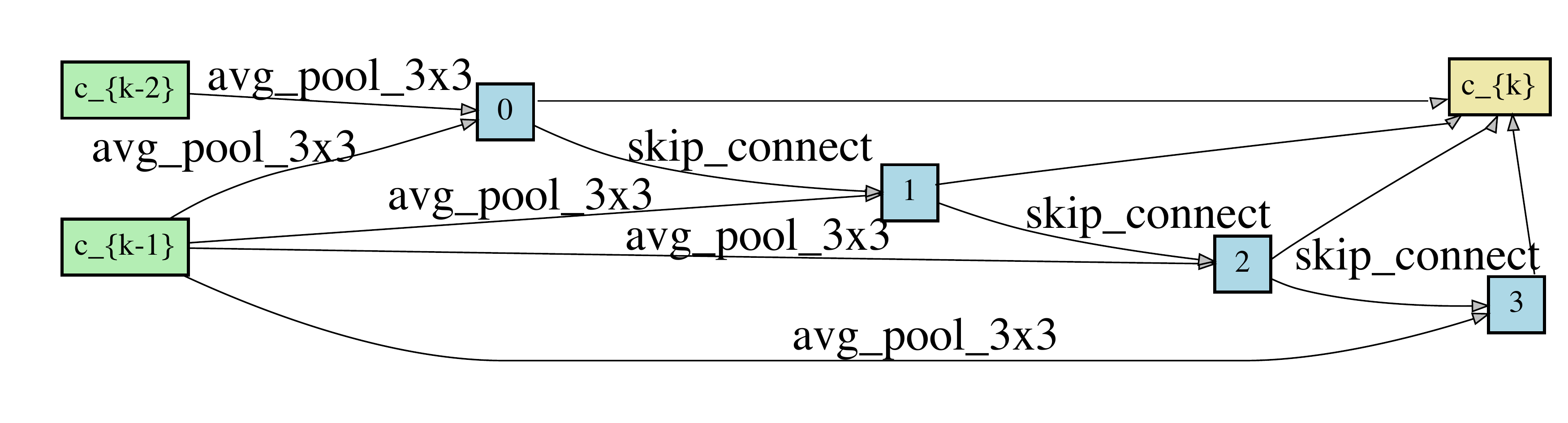}}

       
    \caption{ Architectures found via Vanilla DARTS with 20\% (top) and 50\% (bottom) symmetric noise. Figures (a) and (c) show normal cells and figures (b) and (d) show reduction cells. The reduction cells consist of many parameterless operations.
    }
    \label{fig:vanillaarch}
\end{figure*}

\begin{figure*}[t]
    \centering
     \subfloat[Normal cell (20\%-sym)]{
    \includegraphics[width=0.53\textwidth]{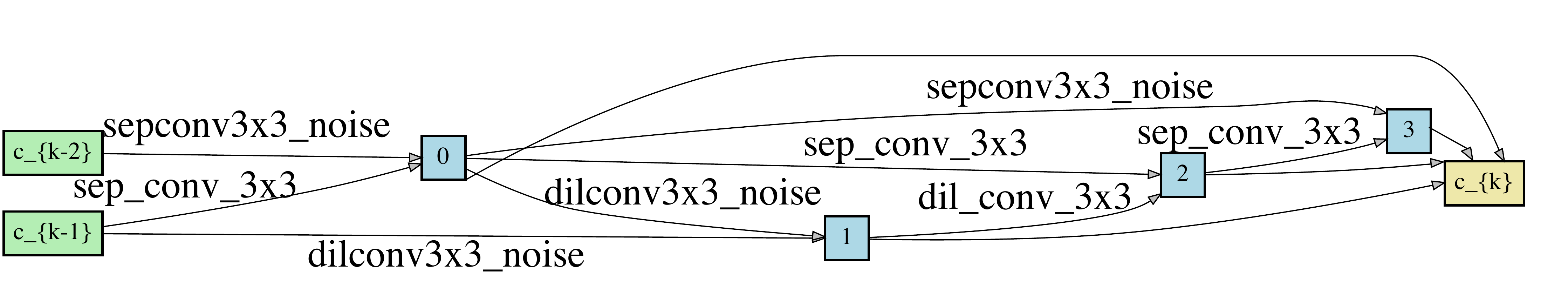}}  
    \subfloat[Reduction cell (20\%-sym)]{
        \includegraphics[width=0.5\textwidth]{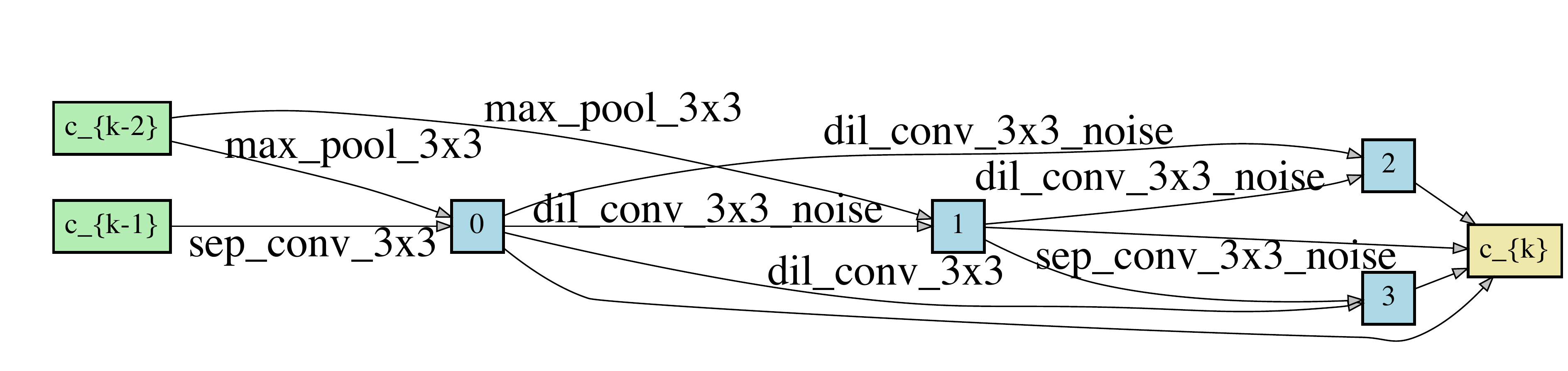}}   
    
     \subfloat[Normal cell (50\%-sym)]{\includegraphics[width=0.55\textwidth]{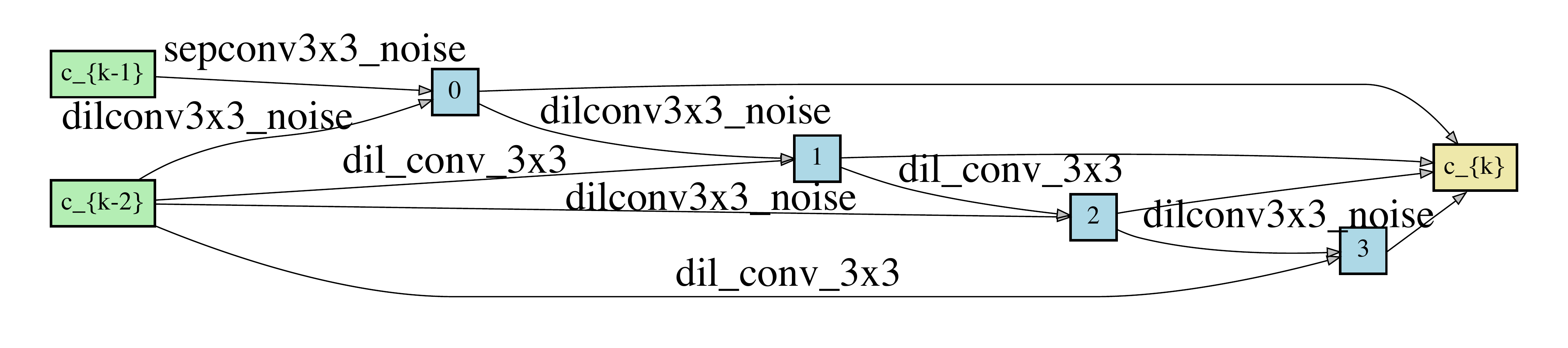}}  
    \subfloat[Reduction cell (50\%-sym)]{
        \includegraphics[width=0.5\textwidth]{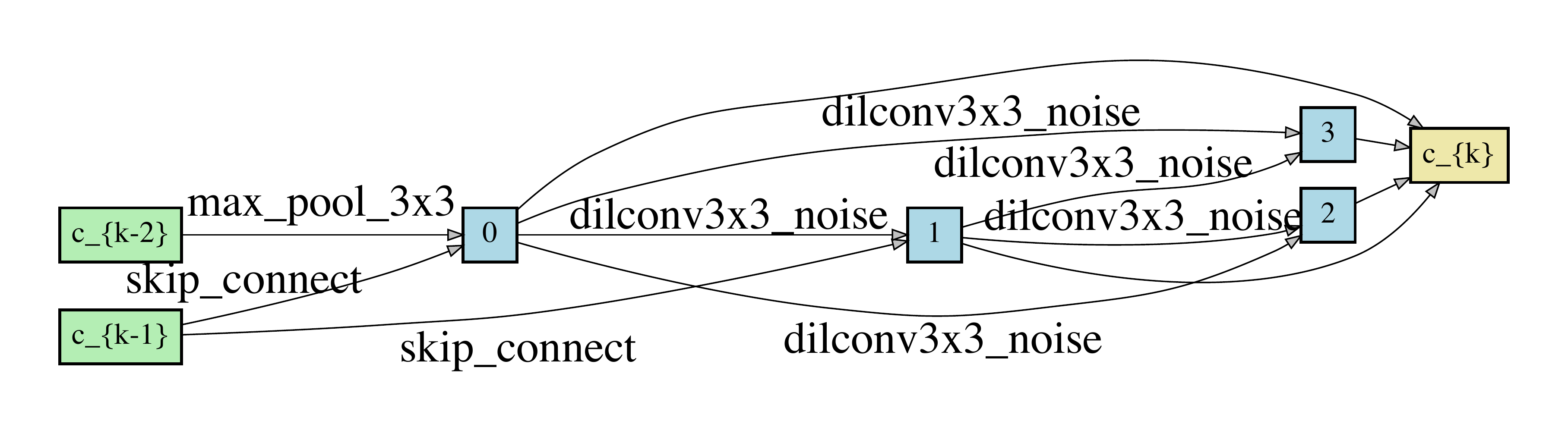}}

       
    \caption{ Architectures found via DARTS + nConv with 20\% (top) and 50\% (bottom) symmetric noise. Figures (a) and (c) are normal cells and figures (b) and (d) are reduction cells. The normal and reduction cells consist of noise convolution.
    }
    \label{fig:nconvarchitecture}
\end{figure*}

In Fig.~\ref{fig:vanillaarch}, we show a  cell architecture  found via Vanilla DARTS with 20\% and 50\% symmetric noise on CIFAR-10. Most of the operations in the reduction cells for both noise cases have no parameters (\eg, skip connection and max-pooling). The parameterless cells in Fig.~\ref{fig:vanillaarch} have a smaller parameter size compared to the cells found with our operator in Fig.~\ref{fig:nconvarchitecture}. For instance, an architecture searched with Vanilla DARTS (16 stacked cells) has 0.6M parameters but an architecture  including nConv (16 stacked cells) has 1.2M parameters.  
It appears that in the presence of the label noise, the standard architecture search engine tries to prevent overfitting by selecting parameterless operations which are not expressive enough and thus lead to sub-optimal architectures with poor performance.

\section{Ablation Study}
\subsection{The impact of the standard deviation}
Below, we perform an ablation study to investigate the impact of different values for standard deviation of the noise injection parameter. The setup for this experiment follows the setup in \textsection~\ref{sec:experiments}  with the toy data from~\cite{shwartz2017opening} contaminated with 20\% symmetric noise and the six layers neural network. We vary the standard deviation in range $\{0.01, 0.05, 0.2, 0.3\}$ and the mean in range $\{0.1, 0.2, 0.3\}$,  and train the model for 600 epochs. 

Fig.~\ref{fig:train_vary_sigma} shows that the standard deviation affects both the accuracy and the training loss when training the model. We observe that training without the noise injection performs poorly because the model overfits to the noisy labels given the lowest training loss attained. We found that injecting noise with standard deviation of 0.2 or 0.3 yields the highest validation accuracy while avoiding overfitting (indicated by the higher and smoothly decreasing training loss). In this experiment, the improvement using the noise injection module yields 18\% improvement over the baseline which does not include the noise injection unit.

We also observe that varying the mean does not affect the performance significantly, thus, we conveniently set the mean to zero for all experiments in this work.

\subsection{Gradient analysis under training with noisy labels}
In this experiment, we  analyze the impact that training samples with clean labels {\em vs.} noisy labels has on the gradient. This helps us understand from which samples the network learns and how overfitting occurs due to the label noise. 

Fig.~\ref{fig:noisegrad} (a) shows the gradient norm as a function of epoch for Vanilla DARTS. The figure shows that in the early epochs, samples with clean labels dominate over samples with noisy labels, whereas in the later epochs of the training, the reverse is true. This phenomenon leads to a performance drop towards the end of training and the model parameters are comparatively more influenced by the noisy samples rather than the clean samples as the network tries to fit to datapoints with erroneous labels.

We then perform the equivalent experiment using our proposed method, shown in Fig.~\ref{fig:noisegrad} (b). The figure  shows that the impact of the gradient of the clean samples dominates the impact of the gradient of the noisy samples throughout the entire training. This illustrates that the risk of overfitting to noisy labels is substantially reduced, and that the network preferentially learns from the clean samples. This observation of how gradient behave translates into the superior performance (under noisy labels) of DARTS + nConv compared to Vanilla DARTS.

\begin{table*}[]
    \centering
    \caption{The Separable Convolution (\textit{SepConv}) and the Separable Noisy Convolution (\textit{SepNConv}).}
    \vspace{0.1cm}
    \begin{tabular}{c c}
    \hline
        Separable Convolution 3x3 (\textit{sep\_conv\_3x3}) & Separable Noisy Convolution 3x3 (\textit{sepconv3x3\_noise})\\
        \hline
           - &\textcolor{red}{Noise injection}\\
          ReLU  & ReLU \\
           3x3 convolution, C channels, no bias &3x3 convolution, C channels, no bias\\
           1x1 convolution, C channels, no bias & 1x1 convolution, C channels, no bias\\
           Batch normalization &Batch normalization\\
            - &\textcolor{red}{Noise injection}\\
          ReLU  &ReLU\\
          3x3 convolution, C channels, no bias  &3x3 convolution, C channels, no bias\\
          1x1 convolution, C channels, no bias &1x1 convolution, C channels, no bias\\
          Batch normalization &Batch normalization\\
    \hline      
    \end{tabular}

    \label{tab:sepconv}
\end{table*}

\begin{table*}[]
    \centering
    \caption{The Dilated Convolution (\textit{DilConv}) and the Dilated Noisy Convolution (\textit{DilNConv}).}
    \vspace{0.1cm}
    \begin{tabular}{c c}
    \hline
        Dilated convolution 3x3 (\textit{dil\_conv\_3x3}) & Dilated Noisy Convolution 3x3 (\textit{dilconv3x3\_noise})\\
        \hline
           - &\textcolor{red}{Noise injection}\\
          ReLU  & ReLU \\
           3x3 convolution, C channels, no bias, dilation 2 &3x3 convolution, C channels, no bias, dilation 2\\
           1x1 convolution, C channels, no bias & 1x1 convolution, C channels, no bias\\
           Batch normalization &Batch normalization\\
    \hline      
    \end{tabular}

    \label{tab:dilconv}
\end{table*}

\end{document}